\documentclass[sigconf]{acmart}

\AtBeginDocument{%
  \providecommand\BibTeX{{%
    \normalfont B\kern-0.5em{\scshape i\kern-0.25em b}\kern-0.8em\TeX}}}

\copyrightyear{2024}
\acmYear{2024}
\setcopyright{acmlicensed}\acmConference[MM '24] {Proceedings of the 32nd ACM International Conference on Multimedia}{October 28--November 1, 2024}{Melbourne, VIC, Australia.}
\acmBooktitle{Proceedings of the 32nd ACM International Conference on Multimedia (MM '24), October 28--November 1, 2024, Melbourne, VIC, Australia}
\acmDOI{10.1145/3664647.3680724}
\acmISBN{979-8-4007-0686-8/24/10}

\usepackage{algorithm}
\usepackage{algpseudocode}
\usepackage{subfig}
\usepackage{hyperref}
\usepackage{multirow}
\usepackage{booktabs}
\usepackage{amsthm,amsmath}
\usepackage{mathrsfs}
\usepackage{graphicx}
\usepackage{color,xcolor}
\usepackage{soul}
\usepackage{enumitem}
\usepackage{algpseudocode}
\usepackage{bbding}
\usepackage{adjustbox}
\definecolor{myorange}{RGB}{197, 90, 17}
\definecolor{mygreen}{RGB}{84, 130, 53}
\definecolor{myred}{rgb}{0.69, 0.25, 0.21}
\definecolor{redbackground}{rgb}{0.99, 0.90, 0.90}
\definecolor{greenbackground}{RGB}{226, 240, 217}
\definecolor{orangebackground}{RGB}{251, 229, 214}
\definecolor{codegreen}{RGB}{0,176,80}
\definecolor{codegray}{rgb}{0.5,0.5,0.5}

\newcommand{\impro}[1]{{\hspace{0.05cm}{\textcolor{red}{\textbf{(+#1)}}}}}

\usepackage{graphicx}

\usepackage{newfloat}
\usepackage{listings}
\DeclareCaptionStyle{ruled}{labelfont=normalfont,labelsep=colon,strut=off} 
\floatstyle{ruled}
\newfloat{listing}{tb}{lst}{}
\floatname{listing}{Listing}

\begin{document}

\title[Edit As You Wish: Video Caption Editing with Multi-grained  User Control]{Edit As You Wish: Video Caption Editing with \\ Multi-grained  User Control}


\settopmatter{authorsperrow=4}

\author{Linli Yao}
\affiliation{%
   \institution{School of Computer Science, \\Peking University}
  \city{Beijing}
  \country{China}}
\email{linliyao@stu.pku.edu.cn}

\author{Yuanmeng Zhang}
\affiliation{%
  \institution{Alibaba Group}
  \city{Beijing}
  \country{China}}
\email{zhangyuanmeng.zym@alibaba-inc.com}

\author{Ziheng Wang}
\affiliation{%
  \institution{School of Information,\\ Renmin University of China}
  \city{Beijing}
  \country{China}}
\email{zihengwang@ruc.edu.cn}

\author{Xinglin Hou}
\affiliation{%
  \institution{Alibaba Group}
  \city{Beijing}
  \country{China}}
\email{xinglin.hxl@alibaba-inc.com}

\author{Tiezheng Ge}
\affiliation{%
  \institution{Alibaba Group}
  \city{Beijing}
  \country{China}}
\email{tiezheng.gtz@alibaba-inc.com}

\author{Yuning Jiang}
\affiliation{%
  \institution{Alibaba Group}
  \city{Beijing}
  \country{China}}
\email{mengzhu.jyn@alibaba-inc.com}

\author{Xu Sun}
\affiliation{%
  \institution{School of Computer Science, \\ Peking University}
  \city{Beijing}
  \country{China}}
\email{xusun@pku.edu.cn}

\author{Qin Jin}
\affiliation{%
  \institution{School of Information, \\ Renmin University of China}
  \city{Beijing}
  \country{China}}
\email{qjin@ruc.edu.cn}
\authornote{Qin Jin is the corresponding author.}

\renewcommand{\shortauthors}{Linli Yao et al.}

\begin{abstract}
Automatically narrating videos in natural language complying with user requests, i.e. Controllable Video Captioning task, can help people manage massive videos with desired intentions.
However, existing works suffer from two shortcomings: 1) the control signal is single-grained which can not satisfy diverse user intentions; 2) 
the video description is generated in a single round which can not be further edited to meet dynamic needs.
In this paper, we propose a novel \textbf{V}ideo \textbf{C}aption \textbf{E}diting \textbf{(VCE)} task to automatically revise an existing video description guided by multi-grained user requests. 
Inspired by human writing-revision habits,  we design the user command as a pivotal triplet \{\textit{operation, position, attribute}\}  to cover diverse user needs from coarse-grained to fine-grained.
To facilitate the VCE task, we \textit{automatically} construct an open-domain benchmark dataset named VATEX-EDIT and \textit{manually} collect an e-commerce dataset called EMMAD-EDIT. We further propose a specialized small-scale model (i.e., OPA) compared with two generalist Large Multi-modal Models to perform an exhaustive analysis of the novel task. For evaluation, we adopt comprehensive metrics considering caption fluency, command-caption consistency, and video-caption alignment. Experiments reveal the task challenges of fine-grained multi-modal semantics understanding and processing. 
Our datasets, codes, and evaluation tools are available at \url{https://github.com/yaolinli/VCE}. 

\end{abstract}





\maketitle

\begin{figure}[t]
    \centering
    \includegraphics[width=0.99\linewidth]{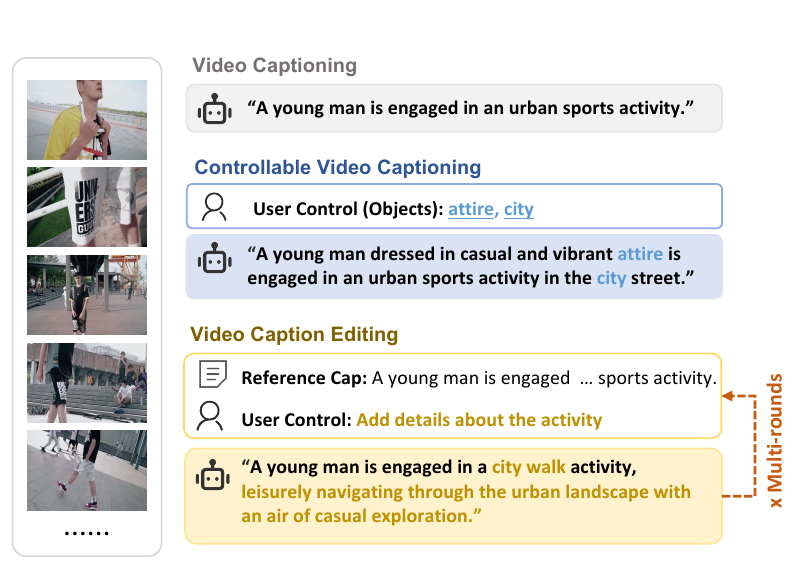}
    \vspace{-8pt}
    \caption{ Comparisons between our proposed Video Caption Editing (VCE) task with conventional video captioning and controllable video captioning. }
    \vspace{-15pt}
    \label{fig:intro}
\end{figure}

\section{Introduction}
The proliferation of videos on the Internet heralds the era of video-dominated media. Video captioning, i.e. automatically describing videos using natural language, has been a prevalent task to assist people in comprehending and managing massive videos.
However, conventional video captioning systems~\cite{7410872, 8099830} tend to generate intention-agnostic descriptions, ignoring the various demands of different users. Therefore, a new task branch, namely controllable video captioning~\cite{chen2019topic,deshpande2018diverse,yuan2020Control,liu-etal-2021-o2na}, has been proposed to integrate user intention as a control signal to guide the description generation.

Although controllable video captioning has great potential in practical applications, existing works have two non-negligible drawbacks. First,
they all employ fixed control signals that can only express \textit{ single-grained controls}, such as Part-of-Speech(POS)~\cite{wang2019controllable}  for structure control, or specified object tags~\cite{liu-etal-2021-o2na} for semantic control. These single-grained controls can not satisfy flexible and diverse user demands.
 Second, they are \textit{single-round controls} that generate a video description once which can not be further revised. Whereas iteratively revising sentences until the ideal texts is a natural process for humans ~\cite{acl2022-iterative}. 
Imagine a real-world scenario, such as E-commerce product promotion, where sellers upload product videos with descriptions to attract customers. There is a good chance that automated video descriptions fail to highlight the seller's preferences. As a result, sellers need to further improve the descriptions by themselves, which is time-consuming and labor-intensive, especially when facing massive long videos. 

We propose a novel \textbf{Video Caption Editing (VCE)} task to automate the video description editing process. The task aims to edit an existing video description conditioned on user commands and video content. As depicted in Figure~\ref{fig:intro}, the inputs of VCE task consist of a video, a reference description, and a user control. It outputs an edited video description based on the user command as a control signal. 
The reference caption can be initialized 
using the last edited output sentence which can thus enable \textit{ multi-round modification}. The VCE task can facilitate personalized video description generation by fulfilling miscellaneous demands from different users or dynamic demands from the same user.

\begin{figure}[t]
    \centering
    \includegraphics[width=0.9\linewidth]{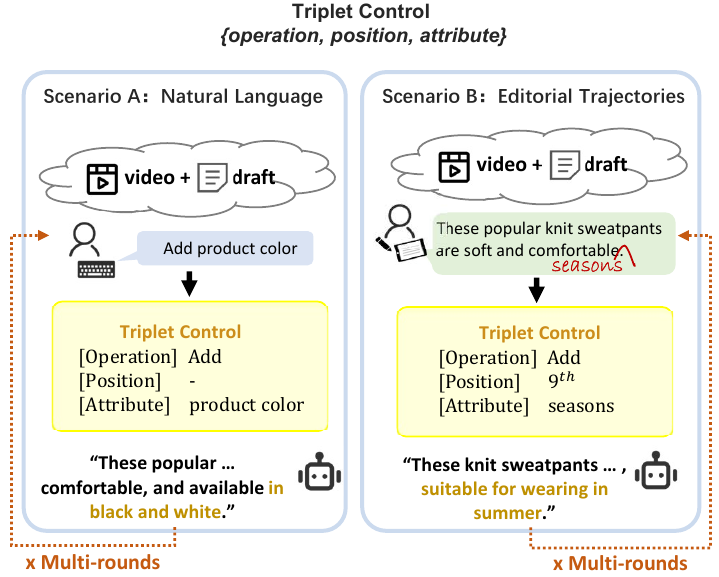}
    \vspace{-8pt}
    \caption{ The triplet control designed in the VCE task can pivot two prevalent interaction signals including natural language (Scenario A) and editing trajectories (Scenario B). }
    \vspace{-10pt}
    \label{fig:senarios}
    
\end{figure}

In the VCE task, how to define the user command to cover multi-grained requests is crucial. 
Inspired by the human writing-revision habits, we unify the user edit commands into a triplet format \{\textit{operation, position, attribute}\} (depicted in Figure~\ref{fig:senarios}) for three 
 advantages. Firstly, it condenses the core elements in an editing operation. Secondly, it can accommodate 
 two prevalent front-end interface signals including natural language 
 and editing trajectories from tablet computers. Finally, the different combinations  
 of three elements in the triplet can cover \textit{ multi-grained} user commands from coarse-grained control (e.g. sentence length change) to fine-grained control (e.g. insert new details in the specific position), as illustrated in Table~\ref{tab:command}.

We collect two novel benchmark datasets named VATEX-EDIT and EMMAD-EDIT to support the exploration of the VCE task. The VATEX-EDIT dataset is automatically constructed from a large-scale video-text dataset VATEX~\cite{wang2019vatex} in the open domain. Meanwhile, to close the gap between research advances and real-life applications, we manually collect an e-commerce editing dataset called EMMAD-EDIT, which is more challenging from two aspects: 1) longer videos (average 27.1 seconds) and longer captions (average $\sim$100 words); 2) external domain knowledge needed to generate product-oriented video descriptions. 
Based on the two benchmark datasets, we propose a specialized model, namely OPA, that converts the command triplet into a textual token sequence to alleviate heterogeneity among multi-grained commands.
We demonstrate the feasibility of utilizing a unified framework to handle seven types of user commands. Moreover, we adopt two generalist Large Multimodal Models (LMMs) as a comparison to gain an in-depth understanding of the characteristics and challenges of the VCE task.

The main contributions of this paper are four-fold. 
1) To the best of our knowledge, we are the first to propose the VCE task to achieve \textit{multi-round} editing and design the user command as a triplet format to express \textit{multi-grained} user requests.
2) We build two benchmark datasets from different domains, including the general domain (VATEX-EDIT) and commercial domain (EMMAD-EDIT), to facilitate the investigation of the VCE task. 
3) We develop an evaluation suite to assess the edited video description based on caption fluency, command-caption consistency, and video-caption alignment.
4) We propose a unified specialist framework OPA and adapt two generalist LMM methods to initially tackle the task, followed by a comprehensive analysis.

\vspace{-5pt}
\section{Related Work}

\begin{table*}[t]

\centering
\caption{ Editing commands via different combination of elements \textit{\{operation, position, attribute}\}. It covers seven multi-grained demands from coarse-grained controls (e.g. expand description) to fine-grained controls (e.g. add specified \textit{attributes} at specified \textit{positions}). The atomic operations consist of \textit{add} and \textit{delete}. The multi-grained commands with a reference caption (e.g. ``A group of girls is on the field playing a game.'') are unified as a control token sequence (Section~\ref{sec:input_foramt_design}) to guide the model. }
\begin{adjustbox}{max width=\linewidth}

\begin{tabular}{@{}ccc|l|l|l@{}}
\toprule
\multicolumn{3}{c|}{\textbf{Command}} & \multirow{2}{*}{\textbf{ \quad  Notation}} & \multirow{2}{*}{\textbf{\quad \qquad \qquad \qquad Demand}}   & \multirow{2}{*}{\textbf{\qquad \qquad \qquad \qquad \qquad Unified Input Control}}    \\  
opera.   & pos   & attr    &                           &                                                                          \\ \toprule
\Checkmark        &         &     &$\langle$\textit{add},\hfill-\hfill,\hfill-\hfill$\rangle$\hfill  & expand description     & \textcolor{myorange}{\colorbox{orangebackground}{[ADD]}} A group of girls is playing a game.                 \\ \midrule

\Checkmark        & \Checkmark         &    &  $\langle$\textit{add}, \textit{pos},\hfill-\hfill$\rangle$\hfill  & expand description at specified \textit{positions}  & \textcolor{myorange}{\colorbox{orangebackground}{[ADD]}} A group of girls is \colorbox{redbackground}{\textcolor{myred}{[MASK]}} playing a game. \\ \midrule

\Checkmark        &        & \Checkmark     & $\langle$\textit{add},\hfill-\hfill, \textit{attr}$\rangle$\hfill    & add specified \textit{attributes} in description   & \textcolor{myorange}{\colorbox{orangebackground}{[ADD]}} \colorbox{greenbackground}{\textcolor{mygreen}{field, hockey;}} A group of girls is playing a game. \\ \midrule

\Checkmark & \Checkmark & \Checkmark &  $\langle$\textit{add}, \textit{pos}, \textit{attr}$\rangle$\hfill & add specified \textit{attributes} at specified \textit{positions}  & \textcolor{myorange}{\colorbox{orangebackground}{[ADD]}} \textcolor{mygreen}{\colorbox{greenbackground}{field, hockey;}} A group of girls is \colorbox{redbackground}{\textcolor{myred}{[MASK]}} playing a game. \\

\midrule
\midrule
\Checkmark        &         &     & $\langle$\textit{del},\hfill-\hfill,\hfill-\hfill$\rangle$\hfill   & shorten description     & \textcolor{myorange}{\colorbox{orangebackground}{[DEL]}} A group of girls is on the field playing a game.                 \\ \midrule

\Checkmark & \Checkmark  & & $\langle$\textit{del}, \textit{pos},\hfill-\hfill$\rangle$\hfill  & shorten description at specified \textit{positions}  & \textcolor{myorange}{\colorbox{orangebackground}{[DEL]}} A group of girls is on \colorbox{redbackground}{\textcolor{myred}{\st{(the filed)} [MASK]}} playing a game. \\ \midrule

\Checkmark        &      &  \Checkmark  & $\langle$\textit{del}, \hfill-\hfill, \textit{attr}$\rangle$\hfill     & delete specified \textit{attributes} from description & \textcolor{myorange}{\colorbox{orangebackground}{[DEL]}} \textcolor{mygreen}{\colorbox{greenbackground}{field, group;}} A group of girls is on the field playing a game. \\ 

\bottomrule

\end{tabular}
\end{adjustbox}

\label{tab:command}
\end{table*}

\noindent\textbf{Controllable Video Captioning.} 
Video captioning~\citep{ramanishka2016multimodal, 7410872, tang2021clip4caption, 8099830,aafaq2019spatio,Luo2020UniVL,pan2020spatio,shetty2016frame,shi2019watch} is a challenging cross-modal task to automatically describe the visual contents of a video in natural languages. In order to satisfy the varied pragmatic interests of different users, controllable video captioning~\citep{chen2019topic,deshpande2018diverse,yuan2020Control,liu-etal-2021-o2na} has been a newly prevalent task. It aims to derive video descriptions conditioned on a predefined control signal, e.g. visual object tags. 
\citet{wang2019controllable} introduce Part-of-Speech(POS) information as guidance and \citet{yuan2020Control} directly utilize an exemplar sentence. Their goal is to generate descriptions with desired syntactic structures. Meanwhile, other works aim to control sentence semantics. \citet{chen2019topic} proposes a topic-guided model to generate topic-oriented descriptions. \citet{liu-etal-2021-o2na} focus on producing object-oriented sentences controlled by multiple user-interested objects. 
However, the above endeavors all generate a sentence once and can't be edited dynamically. Besides, their designed control signals are single-grained which can not cover flexible user intentions. Instead, we define a novel VCE task that can revise a description in multiple rounds covering multi-grained user demands.

\noindent\textbf{Image Caption Editing.} 
Conventional image captioning~\cite{vinyals2015show,xu2015show,liu2016image2text,stefanini2022show,li2016image,liu2018context,Yao2022CapEnrichEC,yang2023visual,yao2022image} generates the description for images from scratch which may lead to factual mistakes. \citet{Sammani2019ModificationNet} firstly define the image caption editing task that modifies an existing caption (a.k.a. reference caption) conditioned on the image content to obtain more accurate descriptions. \citet{Sammani_2020_CVPR} further propose a novel EditNet framework to achieve interactive and adaptive edits. \citet{ATD-Net} design an  adaptive text-denoising network to alleviate the semantic gap between input images and reference sentences.
The above works all edit image captions implicitly. \citet{wang2022explicit} propose the explicit image caption editing task to make the modification process more explainable and efficient. 
In summary, all these works can only correct the wrong content in the reference captions and ignore specific edit intentions of different users. In this paper, we integrate multi-grained user commands into the video description editing process.

\noindent\textbf{Large Multimodal Models.}
Recent months have witnessed the tremendous success of Large Language Models (LLMs)~\cite{Chowdhery-arxiv-2022-PaLM,Ouyang-arxiv-2022-Training,Zeng-arxiv-2022-GLM,Touvron-2023-llama2-arxiv, Touvron-arxiv-2023-LLaMA, chatgptblog, openai2023gpt4} towards artificial general intelligence. Further, Large Multimodal Models (LMMs)~\cite{instructblip,llava,qwen-vl,ye2023mplug, videochat, videollama,  videollava, timechat} endow LLMs with the visual understanding ability by incorporating vision backbones~\cite{vitg, clip,li2022uniformerv2}. 
Existing LMMs primarily bifurcate into two categories: Image Large Language Models (ImgLLMs)~\cite{instructblip,llava,qwen-vl,ye2023mplug,cha2023honeybee} and Video Large Language Models (VidLLMs)~\cite{videochat, videollama, videollava, valley, timechat}. 
The standard architecture of an LMM comprises a vision backbone for encoding images or videos, a projector~\cite{li2023blip2,cha2023honeybee,yao2024deco} to translate visual embeddings into textual semantic space, and an LLM to process all multimodal contexts. 
As the VCE task involves capturing video semantics, understanding textual user controls, and enabling text editing, it can serve as a new touchstone for LMMs.

\section{Video Caption Editing Task}

\subsection{Task Definition}
\label{sec:task_definition}
Given a video $V$ and a reference caption $R = \{r_1, \dots , r_L\}$, the VCE task aims to generate an edited caption $Y = \{y_1, \dots , y_T\}$ according to the user edit command $C$. The edited caption $Y$ should satisfy the constraints of $V$, $R$ and $C$. Given a ground truth  caption $Y^*$, the maximum likelihood estimation (MLE) training objective of VCE task can be formulated as:
\begin{equation}
\mathcal{L}_{\mathrm{MLE}}= -\frac{1}{T} \sum^T_{t=1}\log p\left( y^*_t \mid y^*_{<t}, V, R, C \right)
\end{equation}
The reference caption can be initialized with the output caption from the last round of editing. 
It is also possible to start the editing process using a auto-generated sentence or human-written one as the reference.  
Due to the reference caption input setting, the VCE task can naturally achieve an interactive editing process with successive editing rounds, which is in line with human writing habits \cite{acl2022-iterative}. The interactive and multi-round revisions can help produce descriptions with higher user satisfaction.

\begin{table*}[t]
\centering
\caption{ Data statistics of VATEX-EDIT and EMMAD-EDIT dataset. \# denotes the number. \textit{VTime} refers to the average duration of videos in seconds. \textit{$\text{Len}_{Ref}$} denotes the average length of reference captions and \textit{$\text{Len}_{GT}$} is the average length of ground-truth captions. \textit{Edit Dist} means the average minimum edit distance between reference captions and edited captions.}
\begin{adjustbox}{width=0.92\linewidth}
\begin{tabular}{@{}l|l|rrr|rrr|cccccc}
\midrule
\multirow{2}{*}{\textbf{Dataset}} &
  \multirow{2}{*}{\textbf{Vision}} &
  \multicolumn{3}{c|}{\textbf{\#Videos/Images}} &
  \multicolumn{3}{c|}{\textbf{\#Editing instances}} &
  \multicolumn{1}{l}{\multirow{2}{*}{\textbf{VTime}}} &
  \multicolumn{1}{l}{\multirow{2}{*}{\textbf{$\text{Len}_{Ref}$}}} &
  \multicolumn{1}{l}{\multirow{2}{*}{\textbf{$\text{Len}_{GT}$}}} &
  \multicolumn{1}{l}{\multirow{2}{*}{\textbf{Edit Dist}}}   &
  \multirow{2}{*}{\textbf{Vocab}} \\ 
 &
   &
  \multicolumn{1}{r}{Train} &
  \multicolumn{1}{r}{Val} &
  \multicolumn{1}{r|}{Test} &
  \multicolumn{1}{r}{Train} &
  \multicolumn{1}{r}{Val} &
  \multicolumn{1}{r|}{Test} &
  \multicolumn{1}{l}{} &
  \multicolumn{1}{l}{} &
  \multicolumn{1}{l}{} &
   \\ \midrule
COCO-EE~\cite{wang2022explicit}
&
  Image &
  52,587 &
  3,055 &
  2,948 &
  97,567 &
  5,628 &
  5,366 &
   -    &
  10.3 &
  9.7 &
  10.9 &
  11,802 \\
Flickr30K-EE~\cite{wang2022explicit}
&
  Image &
  29,783 &
  1,000 &
  1,000 &
  108,238 &
  4,898 &
  4,910 &
  - &
  7.3 &
  6.2 &
  8.8 &
  19,124 \\ \midrule \midrule
VATEX-EDIT &
  Video &
  25,467 &
  2,935 &
  5,867 &
  \textbf{784,805} &
  \textbf{91,513} &
  \textbf{181,638} &
  10.0 & 
  \text{14.4} &
  \text{16.0} &
  \text{11.9} &
  \text{21,634} \\
EMMAD-EDIT &
  Video &
  16,176 &
  5,418 &
  5,502 &
  \text{47,569} &
  \text{15,914} &
  \text{16,169} &
  \textbf{27.1} &
  \textbf{91.3} &
  \textbf{93.7} &
  \textbf{17.8} &
  \textbf{44,725} \\ \midrule

\end{tabular}
\end{adjustbox}

\label{tab:data-stat}
\end{table*}

\vspace{-5pt}
\subsection{User Edit Command}
\label{sec:edit_commad}
It is not trivial to define flexible edit commands in the VCE task to meet various realistic user needs. 
We observe that natural language and writing-revision traces are two natural interactive modes. The former can be received from keyboards or speech converters, while the latter conveniently expresses user intentions with the prevalence of tablets and wireless stylus pens
. A command representation compatible with the above two signals is important and meaningful. 
In this paper, we propose a novel command representation in a triplet format  \textbf{\{\textit{operation, position, attribute}\}}, where \textbf{\textit{operations}} control the overall description editing, \textbf{\textit{positions}} specify the editing locations, which could affect the syntax of sentences, and \textbf{\textit{attributes}} guide the editing operation to control the semantic contents of descriptions.

We define the atomic edit operations as \textit{add} and \textit{delete}, considering that the \textit{replace} editing can be decomposed into the two atomic operations (i.e. first \textit{delete} then \textit{add}). Meanwhile, \textit{position} and \textit{attribute} in the triplet are optional, 
therefore, as shown in Table~\ref{tab:command}, seven specific commands\footnote{Note that we omit the command ``\textit{$\langle$del, pos, attr$\rangle$, delete attributes at specified positions}'', as it can be covered by ``\textit{$\langle$del, pos, -$\rangle$}, \textit{delete description at specified positions}''.} via different combinations of \textit{operation, position, and attribute} elements in the triplet can cover multi-grained realistic demands from coarse-grained (global) controls to fine-grained (local) controls. 
 The designed triplet command can be flexibly obtained by processing the inputs from front-end interfaces including natural language and writing-revision traces (details in Appendix E). In the following method and experiments sections, we perform video description editing directly based on the triplet command.

\begin{figure}[tbp]
    \centering
    \includegraphics[width=0.87\linewidth]{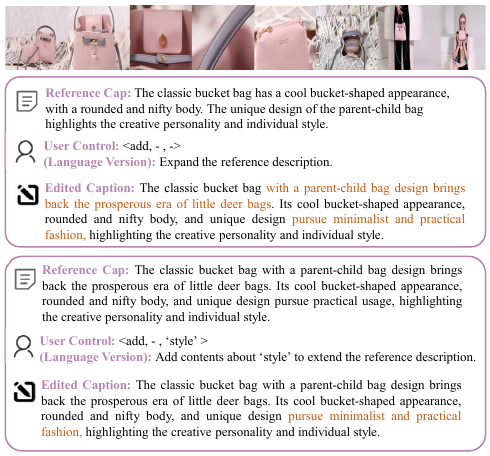}
     \vspace{-8pt}    
    \caption{Annotated data instances of the VCE task.}
    \vspace{-10pt}
    \label{fig:data_case}
\end{figure}

\section{Data Collection}
To faciliate the novel VCE task, we \textit{automatically} construct an open-domain dataset VATEX-EDIT, and \textit{manually} annotate an E-commerce dataset EMMAD-EDIT.
Table~\ref{tab:data-stat} displays the overall data statistics. Compared to prior image caption editing datasets such as COCO-EE and Flickr30K-EE, our new datasets present several distinct advantages: 1) more challenging with the video input and lengthier captions; 2) more diverse encompassing open-domain and e-commerce data; and 3) larger in scale. Specific annotated data instances are illustrated in Figure~\ref{fig:data_case}.

\subsection{VATEX-EDIT Construction}
\label{sec:vatex-data}
It is challenging to construct data samples for the VCE task from scratch, which needs
 a quadruple \textit{(video, command, reference caption, edited caption)} data, abbreviated as (\text{$V, C, R, Y$}). To mitigate the difficulty, we build the VATEX-EDIT dataset by expanding the widely-used video captioning dataset VATEX~\cite{wang2019vatex}, which has high-quality caption annotations for each video. 

 We sample an annotated caption of a video as the reference caption, and the next goal is to construct the \textit{command} and the related \textit{edited caption}.
In general, we aim to construct related \textit{(command, edited caption)} samples including: 
\textbf{1) coarse-grained length-control commands} referring to the global \textit{add} or \textit{delete} edits that result in length changes, and \textbf{2) finer-grained attribute-related commands} to achieve \textit{add} or \textit{delete} attributes.

\begin{figure}[tbp]
    \centering

    \includegraphics[width=0.8\linewidth]{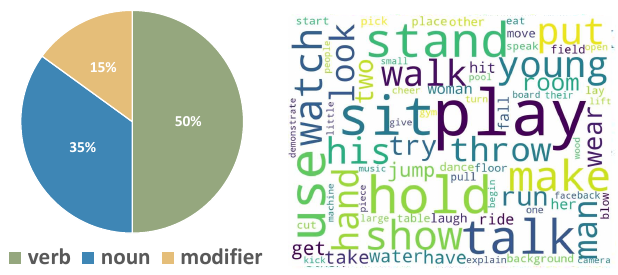}
    \vspace{-8pt}
    \caption{ Attribute statistics on the  VATEX-EDIT.}

    \label{fig:vatex-edit}
\end{figure}

\begin{figure}[tbp]
    \centering
    \vspace{-10pt}
    \includegraphics[width=0.8\linewidth]{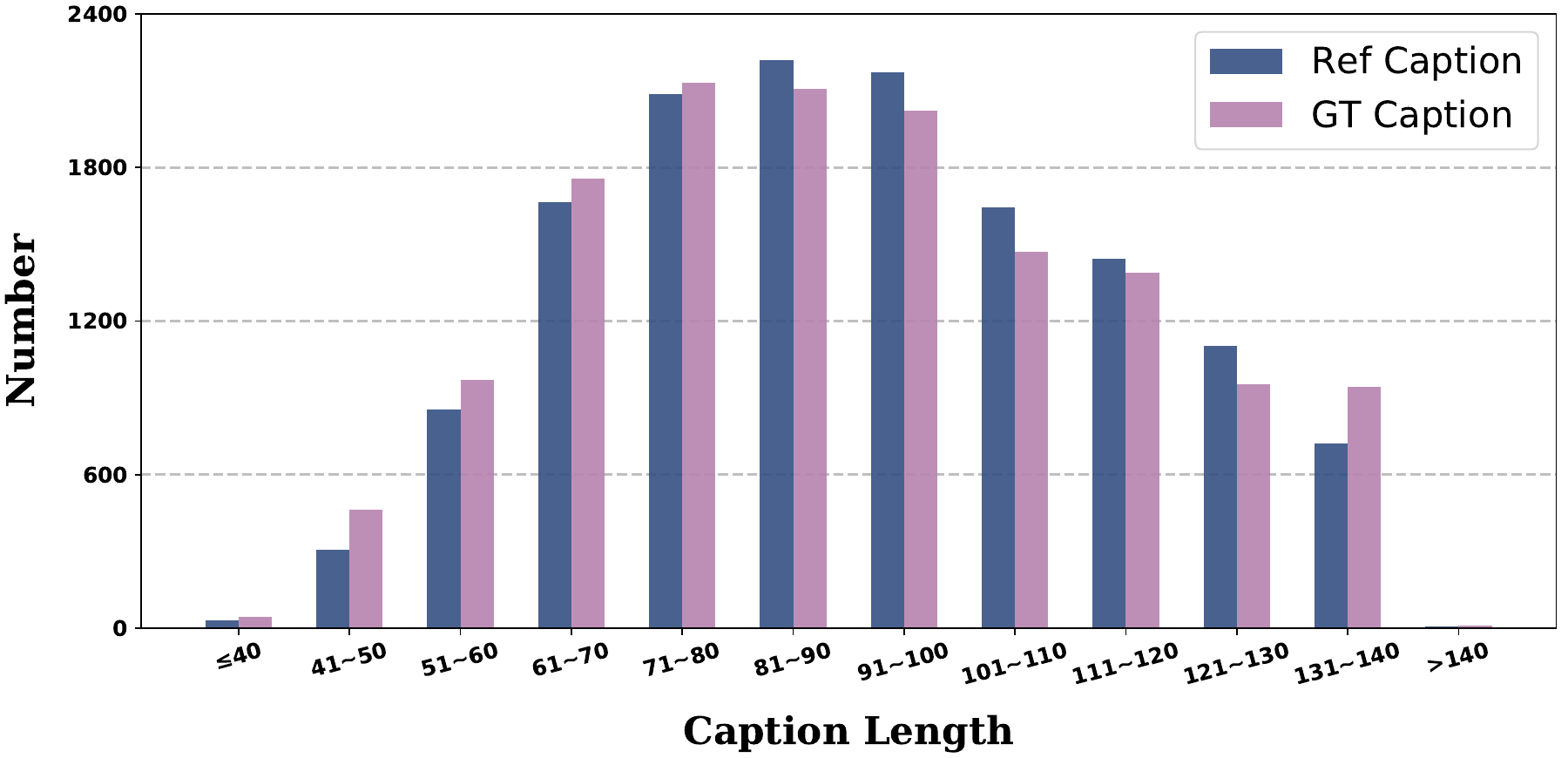}
    \vspace{-10pt}    
    \caption{Caption length distributions on EMMAD-EDIT.}
    \vspace{-10pt}
    \label{fig:emmand-len}
\end{figure}

\textbf{Coarse-grained length-control commands.} 
For \textit{add} operation, we directly select a longer caption with remarkable length differences as the edited caption. The \textit{delete} operation does the exact opposite. Considering the \textit{delete} operation can be easily achieved without video content referring, we replace the reference caption with negative captions that have partially misaligned semantics with the video content. 
Such updated quadruples require models to prioritize removing visually irrelevant content from the reference, which makes the 
\textit{delete}  more challenging.

\textbf{Fine-grained attribute-related commands.}
We construct attribute samples in a ``degradation'' manner, that is firstly detecting attribute words in the reference caption $R$ and then removing them to obtain the edited caption $Y$. We utilize Spacy Syntactic Dependency\footnote{https://spacy.io/} and Semantic Role Labeling~\cite{shi2019semanticRole} to achieve noun, verb or modifier attributes detection and removal in $R$ while maintaining the fluency of $R_{\backslash attr}$ to get $Y$.
After degradation, we can obtain quadruples for the \textit{$\langle$del, -, attr$\rangle$} command.
We exchange the reference caption and the edited caption to obtain the opposite \textit{add} command.
The position information can be naturally recorded to support position-related commands.

\textbf{Statistics and analysis.}
As shown in Table~\ref{tab:data-stat}, compared with existing image 
 editing datasets~\cite{wang2022explicit}, VATEX-EDIT has two salient features: \textit{large-scale} and \textit{diverse}.
Figure~\ref{fig:vatex-edit} visualizes the percentages of modifiers, nouns and verbs in the \textit{attribute}.
Our automatic construction strategy selects verbs as the dominant attributes because verbs are usually related to temporal visual semantics, which is also one of the core challenges of the video description task. The word cloud of specific attribute words shows the \textit{attribute} diversity.

\vspace{-5pt}
\subsection{EMMAD-EDIT Collection}
To satisfy realistic user-demand scenarios, we manually collect a high-quality dataset called EMMAD-EDIT in the E-commerce domain based on a Chinese E-commerce video captioning dataset E-MMAD~\cite{emmad}. The E-MMAD dataset consists of product videos with advertising video descriptions, and additional structure information.
Given a product-oriented video $V$ and an original video description $R$,  we recruit crowd workers to annotate three types of edited descriptions as follows.

\textbf{Simplify original captions } to the target length while maintaining sentence fluency and coherence according to the video content. To ensure the challenge of the VCE task, we require that the length of original sentences should be reduced by at least 20\%.

\textbf{Delete specific attributes.} It aims to select multiple significant attribute words/phrases from $R$ and remove the attribute-related content to get a new caption $Y$. The attributes can be nouns, verbs, or modifiers. To ensure the semantic coherence of $Y$, workers are allowed to modify other parts of $R$ following the ``minimal editing'' principle.

\textbf{ Delete abstract attributes.} We further consider  deleting abstract attributes that do not directly appear in $R$. For example, deleting ``Time and Seasons'' needs to locate season-related content such as ``spring'' and ``summer''. It is more challenging to edit with abstract attributes and also more down-to-earth since user intentions may be vague.

\textbf{Statistics and analysis.}
To ensure annotation quality, extra workers further check the annotated cases. 
Table~\ref{tab:data-stat} shows the specific data statistics. 
EMMAD-EDIT has two remarkable characteristics, i.e.  \textit{long videos} and \textit{long descriptions}. The average video length is 27.1 seconds and the average description length (specified in Figure~\ref{fig:emmand-len}) is around 100 words. We believe the challenging EMMAD-EDIT dataset will promote new technologies for the VCE task.

\begin{figure*}[tbp]
    \centering
    \includegraphics[width=0.98\linewidth]{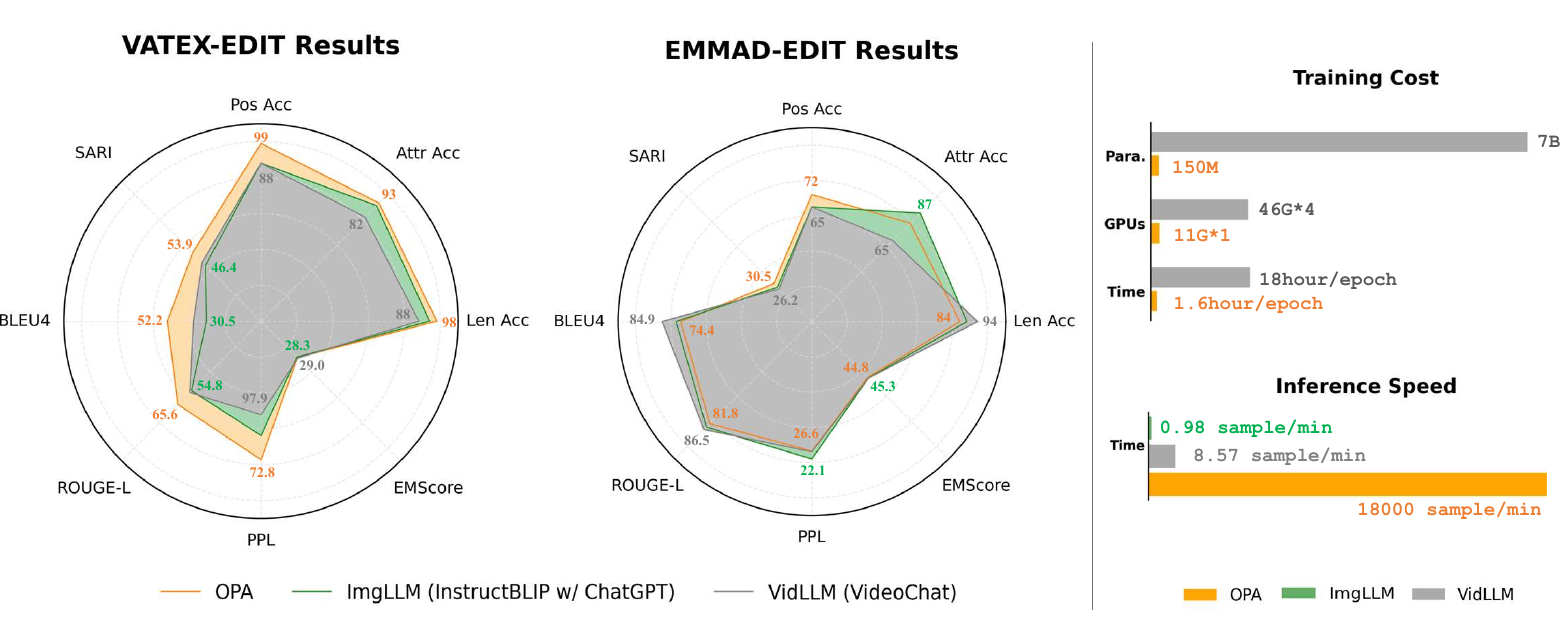}

    \caption{ Overall performance of the small-scale specialist model (i.e., \textcolor{orange}{OPA}) and large-scale generalist models (i.e., \textcolor{green}{ImgLLM} pipeline and end-to-end \textcolor{gray}{ VidLLM}) on the VATEX-EDIT and EMMAD-EDIT dataset. We utilize InstructBLIP~\cite{instructblip} w/ ChatGPT~\cite{chatgptblog} as the ImgLLM pipeline without training. Meanwhile, we conduct instruction tuning on the VideoChat-7B~\cite{videochat} as the end-to-end  VidLLM method.  The training cost (parameters, used GPUs, and training time) and inference speed via a single GPU are on the right. It is noted that we take the negative of PPL$\downarrow$ and rescale it on the radar coordinate for better visualization. }
    \label{fig:main_results}

\end{figure*}

\section{Methodology}
In this section, we begin by introducing how to transform the triplet control into a unified textual sequence.
Subsequently, we explore three approaches for the VCE task to facilitate a comprehensive comparison.
We propose the  \textbf{O}peration-\textbf{P}osition-\textbf{A}ttribute (\textbf{OPA}) model as a small-scale specialist. Additionally, we utilize an Image Large Language Model (ImgLLM) pipeline, and an end-to-end Video Large Language Model (VidLLM) to observe the performance of large multimodal models. Lastly, we develop an evaluation protocol for the novel task.

\subsection{Input Format Design}
\label{sec:input_foramt_design}
We first integrate the seven specific edit commands introduced in Table~\ref{tab:command} into a unified format to achieve multi-grained control.

The main challenge is the heterogeneity among the three elements of command, including \textit{operation}, \textit{position}, and \textit{attribute}. On the one hand, \textit{operations} and \textit{attributes} change the textual semantics while \textit{positions} mainly influence sentence syntax. On the other hand, \textit{attributes} are specific textual words while  \textit{positions} are absolute position indexes. 

To tackle the above challenges, we unify the input format as a textual token sequence. As shown in Table~\ref{tab:command}, we define two special tokens, [ADD] and [DEL], to represent different \textit{add} or \textit{delete} operations. \textit{Attribute} words are naturally text tokens. For \textit{position}, we put special tokens [MASK] in the reference caption to indicate the absolute position indexes. For example, a positioned reference caption ``\textit{A group of girls is} [MASK] \textit{playing a game}'' guides the model to generate new details between words ``is'' and ``playing''. Finally, we concatenate the operation token, the attribute words, and a positioned reference caption as a control sequence to guide the model for description generation. Table~\ref{tab:command} visualizes the input control sequences under seven specific commands when the reference caption is \textit{``A group of girls is  playing a game''.}

\subsection{OPA: A Small-Scale Model as the Specialist}
We construct a small-scale encoder-decoder Transformer architecture, i.e. multi-modal BART~\cite{bart}, to achieve the video description editing task under the guidance of processed control sequences. We utilize the pre-trained BART weights and endow it with the multi-modal ability to understand video content. The specific architecture is depicted in Appendix A.  

\textbf{Input Representation.} Given a video $V$, a reference caption $R = \{r_1, \dots, r_L\}$, and a triplet command $C$, we first extract frame-level visual features and map them to the same dimension as textual embedding. We denote the input attributes as $A = \{a_1, \dots, a_M\}$ and the indicated position index as $l \in [1, L]$. Taking the most fine-grained command ``\textit{add} specified \textit{attributes} at specified \textit{positions}'' as an example, the concatenated control sequence $\widetilde{C}$ 
 for the command is defined as $\{\text{[ADD]}, A, \widetilde{R}\}$. Using  special tokens to separate each part, it is formulated as:
 \begin{equation}
    \widetilde{C} = \{\text{[opera]\;[ADD]\;[/opera]} \  \text{[attr]}\;A\;\text{[/attr]} \  \text{[ref]}\;\widetilde{R} \;\text{[/ref]} \}
\end{equation}
where the positioned reference caption $\widetilde{R}$ is formulated as:
\begin{equation}
    \widetilde{R} = \{ r_1, \dots, r_{l-1}, \text{[MASK]}, r_{l+1}, \dots, r_L \}   
\end{equation}

Finally, we input the visual features $\{V_1, \dots, V_N\}$ and the textual control sequence embedding $W_{\widetilde{C}} = \{W_{\text{[o]}}, W_{\text{[ADD]}}, \dots, W_{\text{[/r]}}\}$ to the Transformer encoder. If the \textit{position} is empty in the command, we input the original reference caption $R$. When the \textit{attribute} is empty in the command, we set $A$ as an empty set.

\textbf{Leverage Pre-trained Knowledge.} The overall training objective as formulated in Section~\ref{sec:task_definition} is to generate an edited description conditioned on the video features and the control sequence. It is worth noting that we keep the [MASK] token consistent with the same token in the \textit{Text Infilling} pre-trained task of BART to leverage the intrinsic pre-training textual ability.

\subsection{LMMs as Contrastive Generalists}
Large multimodal models integrate the advantages of visual understanding and remarkable natural language processing abilities (e.g., text editing) from LLMs. It is significant to probe their performance on the VCE task. Consequently, we explore two typical branches of LMMs including an ImgLLM pipeline and an end-to-end VidLLM.

\textbf{ImgLLM Pipeline.}
We utilize the InstructBLIP~\cite{instructblip} as the ImgLLM. Nevertheless, ImgLLM can only handle images or a single video frame as visual input. To adapt the ImgLLM to the VCE task, we combine InstructBLIP with 
ChatGPT~\cite{chatgptblog}. In this way, InstructBLIP transforms visual semantics at the frame level into textual context, while ChatGPT consolidates all textual task context and achieves caption editing. Specifically, we extract frames from a given video and utilize InstructBLIP to produce detailed visual descriptions for each frame. The frame descriptions with the VCE task definition, task guidelines, and in-context demonstrations~\cite{dong2022survey} of the relative command type will be combined as the final prompt to the ChatGPT.  Instruction details are provided in Appendix A.

\textbf{End-to-end VidLLM.} As VidLLM can handle video tasks directly, we employ the 
VideoChat~\cite{videochat} as an end-to-end LMM solution.
Specifically, we reformat the VATEX-EDIT and EMMAD-EDIT datasets into question-answer chat samples (refer to Appendix A for specifics) and conduct further instruct-tuning on VideoChat-7B using two datasets respectively.

\vspace{-12pt}
\subsection{Evaluation Suite}
\label{sec:evaluation}
How to evaluate the novel VCE task is another noteworthy challenge. Conventional video captioning tasks adopt widely-used captioning metrics such as BLEU4, METEOR, and CIDEr. However, these reference-based metrics only measure the consistency between generated captions and ground-truth annotations, which are insufficient. In this paper, we evaluate the VCE task from three aspects:
1) fluency, 2) controllability, and 3) text-video alignment.

\textit{Fluency.}
Following the previous work~\cite{wang2022explicit}, we adopt widely-used \textbf{BLEU4}~\cite{bleu} and \textbf{ROUGE-L}~\cite{rouge} metrics to measure the overall generation quality. We also use the \textbf{Perplexity (PPL)}~\cite{jelinek1977ppl} metric that reflects the grammatical correctness and semantic meaningfulness. 

\begin{table*}[t]
\centering
\caption{ Ablation study of the OPA model on the VATEX-EDIT dataset. \textit{Multimodal BART} is the backbone of OPA framework. \textit{Pure Transformer} is the same model without pre-trained BART parameters. \textit{Vision Align} means the vision-text alignment.}

\begin{tabular}{l|l|cccc|ccc|c}
\midrule
\multirow{2}{*}{} & \multirow{2}{*}{\textbf{Model}} &  \multicolumn{4}{c|}{\text{\textit{Controllability}}} & \multicolumn{3}{c|}{\text{\textit{Fluency}}} & \text{\textit{Vision Align}} \\
                   &   & \textbf{Len-Acc} & \textbf{Attr-Acc} & \textbf{Pos-Acc} &  \textbf{SARI} & \textbf{BLEU4} & \textbf{ROUGE-L} & \textbf{PPL$\downarrow$} &  \textbf{EMScore} \\ \midrule
1& \text{$\text{Multimodal BART}$}           & -    & -       & -    & 49.7       & 48.0        & 62.0      & 73.9   & 28.7       \\
2& \text{${\text{Multimodal BART}}_{Opera.}$}            & 97  & -   & - & 52.1    & 49.6     & 63.2    & 70.6 & 28.7  \\ 
3& \text{${\text{Multimodal BART}}_{Opera.+Attr}$}       & 97     & 93      & -    & 53.8       & 52.3        & 65.7       & 72.7    & 28.7       \\
4& $\text{OPA}$   & 98     & 93       & 99  &53.9   & 52.2    & 65.6       & 72.8        & 28.7      \\ \midrule
5& \text{${\text{Pure Transformer}}_{Opera.+Pos+Attr}$}     & 98     &  82       & 97    & 52.6      & 50.5       & 64.4      & 77.2   & 28.7      \\ 
6& $\text{3 Single-grained Models}$     & 96    & 69    & 99    & 53.3  & 51.5  & 64.6    & 72.8  &  28.6    \\ \midrule
\end{tabular}
\label{tab:vatex_all}
\end{table*}

\begin{table*}[t]
\centering
\caption{ Overall and breakdown performances on the EMMAD-EDIT dataset.}
\begin{adjustbox}{max width=\linewidth}

\begin{tabular}{l|c|cccc|ccc|c}
\midrule
\multirow{2}{*}{} & \multirow{2}{*}{\textbf{Command}} &  \multicolumn{4}{c|}{\text{\textit{Controllability}}} & \multicolumn{3}{c|}{\text{\textit{Fluency}}} & \text{\textit{Vision Align}} \\
                   &   & \textbf{Len-Acc} & \textbf{Attr-Acc} & \textbf{Pos-Acc} &  \textbf{SARI} & \textbf{BLEU4} & \textbf{ROUGE-L} & \textbf{PPL$\downarrow$} &  \textbf{EMScore} \\ \midrule
1 & $\langle$\textit{add},\hfill-\hfill,\hfill-\hfill$\rangle$\hfill          & 57  & -  & -  & 26.2 & 62.1 & 73.9 & 23.7 & 44.7 \\

2 &  $\langle$\textit{add}, \textit{pos},\hfill-\hfill$\rangle$\hfill    & 75  & -  & 59 & 27.0 & 83.6 & 90.3 & 25.1 & 44.9 \\

3 &  $\langle$\textit{add},\hfill-\hfill, \textit{attr}$\rangle$\hfill   & 80  & 74 & -  & 31.7 & 84.7 & 89.9 & 26.6 & 45.2 \\ 

4 &  $\langle$\textit{add}, \textit{pos}, \textit{attr}$\rangle$\hfill  & 92  & 70 & 85 & 32.9 & 88.1 & 93.3 & 25.5 & 44.8 \\ \midrule

5 & $\langle$\textit{del},\hfill-\hfill,\hfill-\hfill$\rangle$\hfill  & 100 & -  & -  & 33.5 & 66.8 & 73.8 & 30.5 & 44.6 \\

6 & $\langle$\textit{del}, \textit{pos},\hfill-\hfill$\rangle$\hfill   & 99  & -  & -  & 30.7 & 83.9 & 90.6 & 28.8 & 44.6 \\
7 & $\langle$\textit{del}, \hfill-\hfill, \textit{attr}$\rangle$\hfill & 100 & 93 & -  & 33.6 & 75.2 & 83.6 & 28.9 & 44.7 \\
 \midrule

8 & Overall           & 84  & 79 & 72 & 30.5 & 74.4 & 81.8 & 26.6 & 44.8 \\ \midrule
\end{tabular}
\end{adjustbox}

\label{tab:emmad_breakdown}
\end{table*}

\textit{Controllability.} 
Measuring whether an edited caption strictly follows control signals is important for the VCE task. Inspired by previous work~\cite{acl2022-iterative} 
, we first utilize the \textbf{SARI}~\cite{sari} metric to measure the overall edit quality, i.e. the consistency between expected-to-edit and actually-edited spans. 
Moreover, we design three breakdown metrics namely
 \textbf{Length Accuracy, Attribute Accuracy} and \textbf{Position Accuracy} to measure whether the edited caption satisfies the \{\textit{operation, position, attribute}\} triplet control. Concretely, Len-Acc reflects the length change accuracy.
 Attr-Acc checks the appearance of commanded attribute words. 
 Pos-Acc evaluates whether the model inserts/removes content in the specified positions.

 \textit{Text-Video alignment.} The VCE task inherently requires the alignment between edited descriptions and the given video. We use \textbf{EMScore}~\cite{EMScore} to calculate the semantic similarity between edited captions and videos. It focuses on both coarse-grained similarity (video-sentence) and fine-grained similarity (frame-word).

\section{Experiments}

\subsection{Implementation Details}
We implement the small-scale OPA model based on Huggingface Transformers library~\cite{wolf2020transformers}. The default setting is initialized by the $\text{BART}_{base}$. 
We get video frames using fps=1. 
For the VATEX-EDIT dataset in \textit{English}, we adopt BLIP~\cite{li2022blip} ViT-B/16 to extract frame features. The max frame sequence N is set to 20. 
For the EMMAD-EDIT dataset in \textit{Chinese}, we initialize our model with the Chinese $\text{BART}$.
We adopt CN-CLIP~\cite{chinese-clip} ViT-B-16 to extract video frame-level features. The max frame sequence N is set to 30 and the max decoding length is set to 150. 
For training, we use AdamW~\cite{adamw} with a learning rate of 1e-5 and optimize for 20 epochs with a batch size of 20. During inference, we set the beam size 
 of generation as 5.
The ImgLLM pipeline utilizes the identical frame number N as the OPA model. This pipeline doesn't involve any training. We choose one in-context learning sample for every command type integrated into the ChatGPT prompt. For VideoChat model, we set frame number N as 8 to fit its default setting. We conduct further instruct-tuning on the official 7B checkpoints with batch size 64.

\begin{figure*}[t]
    \centering
    \includegraphics[width=0.95\linewidth]{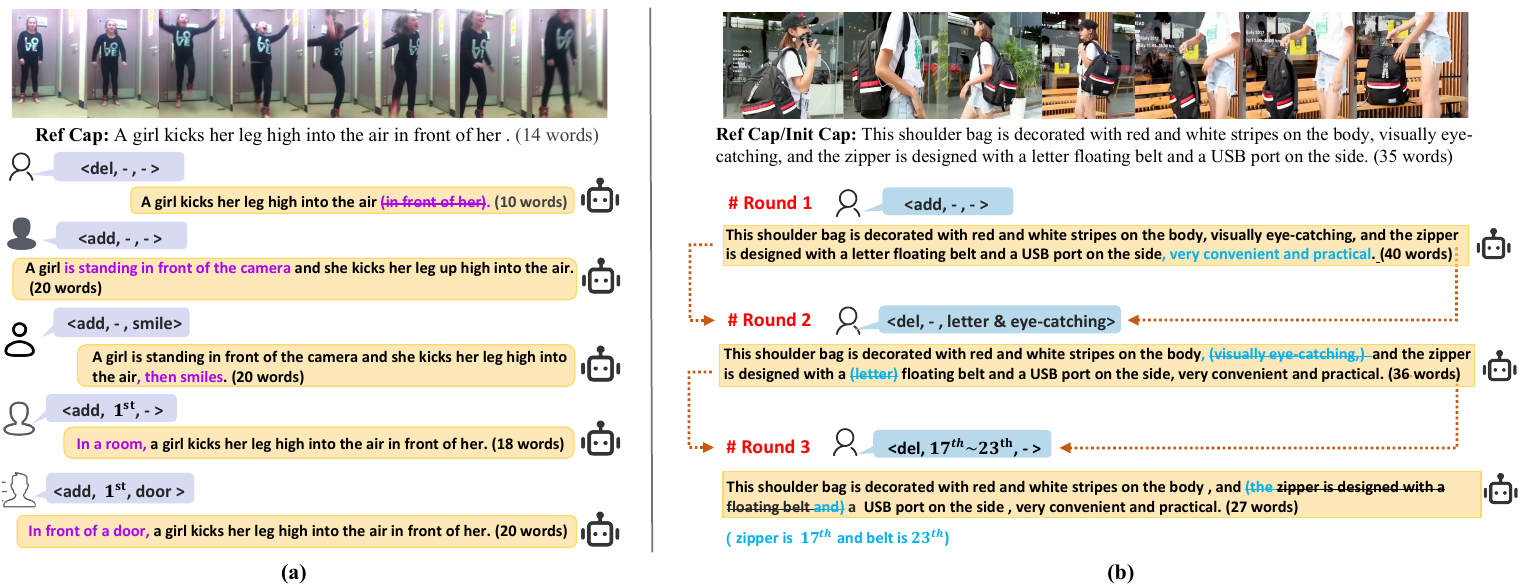}

    \caption{ Visualization of \textbf{(a) multi-grained command editing} and (b) successive multi-round editing using the OPA model.}
    \label{fig:case_study}

\end{figure*}

\vspace{-5pt}
\subsection{Compare Specialist and Generalist Models}
Figure~\ref{fig:main_results} shows the overall performance of three baselines on the VATEX-EDIT and EMMAD-EDIT datasets. 
Interestingly, overall performances are divergent across the two datasets. On the large-scale open-domain VATEX-EDIT dataset, the small-scale specialist OPA model with only 150M parameters outperforms the LMM approaches. It suggests that with sufficient training instances (784,805 samples in VATEX-EDIT), a small specialized model has the potential to perform more effectively and efficiently. 

On the e-commerce EMMAD-EDIT dataset, the LMM methods achieve higher scores across most metrics. 
EMMAD-EDIT is more challenging because it requires domain knowledge, such as unseen product attributes and advertising description style, and has limited training data (refer to Table~\ref{tab:data-stat}). Results show that ImgLLM with InstructBLIP and ChatGPT achieve highest Attr-Acc (87\%) even without training. We argue that on this domain, generalist methods are more promising to leverage their intrinsic knowledge to edit product-related descriptions. 

Despite performance advantages, the training cost and inference speed must be taken into account due to the booming number of videos. As illustrated in Figure~\ref{fig:main_results} (right), compared to the VidLLM and ImgLLM, the OPA model demonstrates significant benefits over both VidLLM and ImgLLM in terms of lower training costs and faster inference speeds. Designing a model that balances performance with speed and cost represents a crucial trade-off. 

Although both specialist and generalist models offer unique advantages, there remains considerable scope for further enhancements to develop an effective editing system, especially on the EMMAD-EDIT dataset. The \textit{Controllability} (Pos-Acc 72\%, Attr-Acc 87\% and Len-Acc 94\%) on the EMMAD-EDIT dataset is insufficient. Moreover, the alignment between video and caption, as indicated by the EMScore, requires significant improvement. In conclusion, the key of the VCE task lies in the combination of fine-grained video and user control understanding and precise text editing capabilities.

\begin{table}[tbp]
\small
\centering
\caption{ The effects of visual modality on the VATEX-EDIT.}
\vspace{-8pt}
\begin{adjustbox}{max width=\linewidth}
\begin{tabular}{c|c|lll}
\midrule
\textbf{Command}                         & \textbf{Video} & \textbf{EMScore} & \textbf{SARI} & \textbf{BLEU4}  \\ \midrule
\multirow{2}{*}{Overall}         & \XSolidBrush     &  28.1    &  53.5    & 51.7           \\
                                & \CheckmarkBold     &  28.7 \impro{0.6}      & 53.9 \impro{0.4}     & 52.2\impro{0.5}            \\ \midrule
\multirow{2}{*}{ $\langle$\textit{del},\hfill-\hfill,\hfill-\hfill$\rangle$\hfill}         & \XSolidBrush     &  27.0    &  39.4    & 12.1         \\
                                & \CheckmarkBold     &  28.5 \impro{1.5}      & 40.4  \impro{1.0}     & 13.5\impro{1.4}            \\ \midrule
\multirow{2}{*}{$\langle$\textit{add},\hfill-\hfill,\hfill-\hfill$\rangle$\hfill  }        & \XSolidBrush        & 28.3        &  41.9    & 10.6         \\
                                &  \CheckmarkBold      &29.0 \impro{0.7}         & 42.0  \impro{0.1}     & 11.1 \impro{0.5}        \\ 
                                \midrule
\end{tabular}
\end{adjustbox}
\vspace{-5pt}

\label{tab:vatex_vision}
\vspace{-8pt}
\end{table}

\vspace{-7pt}
\subsection{Further Task Analysis}
We conduct further ablation studies on the small-scale OPA model to delve into a detailed analysis of the VCE task.

\noindent\textbf{Increase control signals.}
We analyze the editing performance under different control signals in Table~\ref{tab:vatex_all}. The proposed OPA framework achieves high \textit{controllability} accuracy (Len-Acc 98\%, Attr-Acc 93\%, and Pos-Acc 99\%) while maintaining sentence quality. 
The first block (lines 1-4) shows the controllability accuracy and caption quality when progressively inputting more control signals into the model. With the increasing aspects of control signals, there is no decline in sentence fluency and text-vision alignment. It indicates that our model can edit the reference caption with reasonable syntactic and semantic changes under multi-aspect guidance.
Line 5 shows the result of Pure Transformer trained from scratch. Without BART pre-trained parameters, the overall controllability and fluency metrics decrease (SARI from 53.9 to 52.6, BLEU4 from 52.2 to 50.5), which verifies the benefits of textual pre-training knowledge.

\noindent\textbf{Unified framework vs separate models.} 
To satisfy different-granularity commands, we compare the performances of training a unified OPA model vs. training multiple separate models in Table~\ref{tab:vatex_all}. 
In the \textit{3 Single-grained Models} setting (line 6), we train three models respectively to deal with three control granularities, i.e. \{\textit{operation}\}, \{\textit{operation, attribute}\}, and \{\textit{operation, position, attribute}\}.
The OPA model reaches a remarkably higher score on the Attr-Acc (93\% vs 69\%) with better SARI, BLEU4, and ROUGE-L. It demonstrates that our unified input design can alleviate the confusion and heterogeneity of multi-grained commands.

\begin{table}[t]
\small
\centering
\caption{ Mean score (rated 1-5) of the human evaluation on the two datasets. \textit{Trans.} is short for Pure Transformer.}
\vspace{-8pt}
\begin{adjustbox}{max width=\linewidth}
\begin{tabular}{@{}l|l|ccc@{}}
\midrule
\textbf{Dataset} & \textbf{Model} &  \textit{Control.} & \textit{Fluency} & \textit{Vision Align}  \\ \midrule
\multirow{3}{*}{EMMAD-EDIT} & Trans.   & 2.94   &  3.02    &  3.27   \\
& OPA  & 3.98     &  3.85     &  3.76    \\ 
& GT  & 4.67     &  4.37    &  4.22    \\ 
\midrule 
\multirow{3}{*}{VATEX-EDIT} & Trans.    & 4.18     &  4.23     &  3.76   \\
& OPA  & 4.36     & 4.43      &  3.93    \\ 
& GT  &  4.48    & 4.34      & 4.41     \\ 

 \midrule
\end{tabular}
\end{adjustbox}
\vspace{-8pt}

\label{tab:human_eval}
\vspace{-5pt}
\end{table}

\noindent\textbf{Difficulty level of various commands.} 
Table~\ref{tab:emmad_breakdown} (lines 1-7) displays the performances of different commands, indicating their respective difficulty levels. 
Generally, the \textit{add} operation proves to be more challenging than \textit{delete}, primarily because it requires the constraint of video content. For the \textit{add} operations, the finer the command granularity (lines 1-4), the higher the \textit{controllable} and \textit{fluency} scores. It reveals that when provided with more detailed control signals, the model can generate desired captions more easily.

\noindent\textbf{Effects of vision modality.} 
We compare the model performance with and without video input in Table~\ref{tab:vatex_vision}. Adding vision modality brings overall metric improvements since it provides visual semantics to guide edited video description generation. 
For \textit{$\langle$del,-,-$\rangle$} command, we especially construct challenging samples in which reference captions have misalignments with videos (Section~\ref{sec:vatex-data}). With the visual semantics, our model prioritizes removing the video-misalignment contents and achieving a higher EMScore (from 27.0 to 28.5).
Similarly, \textit{$\langle$add,-,-$\rangle$} command requires enriching the original caption referring to the video content.

\subsection{Quantitative Results}

\noindent\textbf{Multi-grained editing controls.}
Provided with various commands, the OPA model can 
output different edited descriptions to satisfy multi-grained user requests. As Figure~\ref{fig:case_study} (a) shows, our OPA model successfully generates different desired descriptions given the same video, the same reference caption but different commands from coarse-grained (e.g. \textit{$\langle$add, -, -$\rangle$}) to fine-grained (e.g. \textit{$\langle$add, $1^{st}$, door$\rangle$}).

\noindent\textbf{Successive editing controls.}
The OPA model also supports interactive editing with successive controls in the VCE task, depicted in Figure~\ref{fig:case_study} (b). The edited description can serve as the reference caption in the next round to make further editing to satisfy dynamic user demands.

\noindent\textbf{Human Evaluation.}
We further adopt human evaluation to assess the quality of edited video descriptions. We recruit 20 evaluators to score the generated descriptions. We randomly sample 200 test cases from VATEX-EDIT and 350 cases from EMMAD-EDIT respectively. During the evaluation, we randomly order the edited captions generated from \textit{Pure Transformer baseline}, \textit{OPA}, and \textit{groundtruths (GT)}. The evaluators are asked to rate each description from three aspects on a scale of 1 to 5 points. Table~
\ref{tab:human_eval} shows the OPA model exceeds the controllable Transformer baseline in three aspects, especially the \textit{controllability}.

\section{Conclusion}
We propose a novel multi-modal task named Video Caption Editing (VCE), which aims to automatically edit video descriptions under the guidance of multi-grained user commands. To satisfy diverse and varied user demands, we design the user control signal as a \{\textit{operation, position, attribute}\} triplet to flexibly cover both coarse-grained and fine-grained controls. We collect two datasets named VATEX-EDIT and EMMAD-EDIT from different domains. We further employ comprehensive metrics to assess fluency, controllability, and vision-text alignment. Finally, we introduce a small specialized model called OPA, an ImgLLM pipeline, and an end-to-end VidLLM to dive into the task challenges and provide good starting points.

\textbf{Limitations and Future Work.}
This paper primarily introduces appropriate baseline solutions for the VCE task, aiming to provide a thorough analysis. Nonetheless, it falls short of designing architectural innovations, leaving ample room for exploration in the future. Further insights into significant future directions are discussed in Appendix F.


\begin{acks}
This work was partially supported by the  National Natural Science Foundation of China (No. 62072462), the Fundamental Research Funds for the Central Universities, and the National Natural Science Foundation of China (No. 62176002). 
\end{acks}

\bibliographystyle{ACM-Reference-Format}
\bibliography{sample-base}

\clearpage

\appendix

\noindent In the supplementary material, we first introduce details of three approaches (Section ~\ref{sec:chatgpt}), then present more construction details about the VATEX-EDIT dataset (Section~\ref{sec:vatex-edit}) and EMMAD-EDIT dataset (Section~\ref{sec:emmad-edit}). Moreover, we provide the pseudo-code of the Position Accuracy metric (Section~\ref{sec:metric}) and the conversion instructions from interface signals to triplet format (Section~\ref{sec:conversion}). Finally, based on a good start on the task, dataset and method foundation, we further discuss a variety of interesting aspects worth exploring in the future (Section~\ref{sec:future}) and the related social impact (Section~\ref{sec:social}).

\section{Model Details}
\label{sec:chatgpt}
\subsection{Architecture of OPA Model}

\begin{figure}[hbp]
    \centering
    \includegraphics[width=0.95\linewidth]{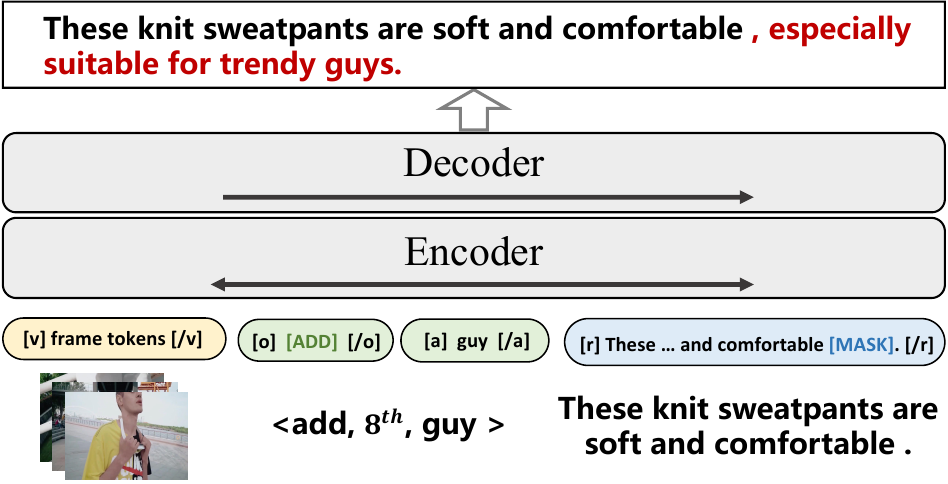}
    \caption{The overall \textbf{OPA} framework. $8^{th}$ denotes the specified position of the $8^{th}$ word.}
    \label{fig:model}
\end{figure}

The overall architecture of the specialist model OPA is depicted in Figure~\ref{fig:model}, which is built on the BART model.

\begin{figure}[tbp]
    \centering
    \includegraphics[width=\linewidth]{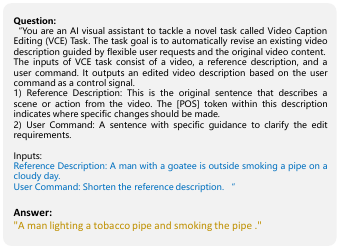}
    \caption{A question-answer chat example used to conduct instruction tuning on the VidLLM.}
    \label{fig:videochat-train-sample}
\end{figure}

\begin{figure*}[tbp]
    \centering
    \includegraphics[width=\linewidth]{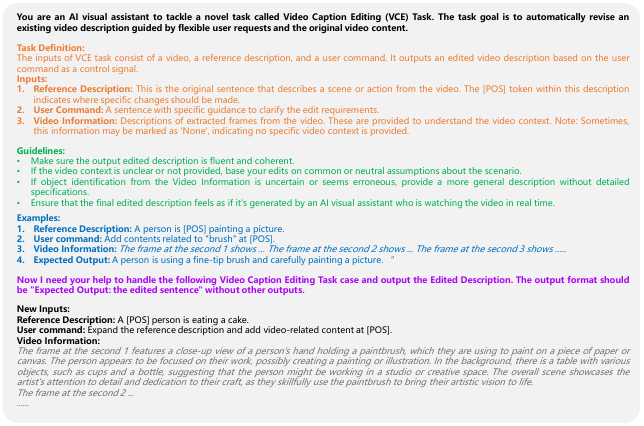}
    \caption{Designed prompts for the ImgLLM pipeline. It encompasses the task definition, helpful guidelines, an in-context learning~\cite{dong2022survey} example, exhaustive visual descriptions, and the new case to be solved. With the well-designed prompt, ChatGPT can output the desired edited description with the required format. We manually construct seven in-context demonstrations involving each specific command type. For each input prompt, the specific command type of the in-context demonstration is aligned with the new case to help ChatGPT better understand the task.}
    \label{fig:chatgptprompt}
\end{figure*}

\subsection{Prompt of ImgLLM Pipeline}

We build a ChatGPT pipeline with the visual expert model, i.e. InstructBLIP, to verify the vision-enhanced LLM performance on the VCE task.
Specifically, we extract key frames from a given video and utilize InstructBLIP to produce detailed descriptions for each frame. We uniformly select 20 frames for VATEX-EDIT dataset and 30 frames for EMMAD-EDIT dataset, which is exactly the same frame number that our OPA model uses. These frame descriptions with timestamps help the ChatGPT to understand the exhaustive video content.

\noindent\textbf{Designed Prompts.}
As Figure~\ref{fig:chatgptprompt} illustrates, we conduct a well-designed prompt for ChatGPT including the task definition, 
helpful guidelines, an in-context learning~\cite{dong2022survey} example, the video information, and the new case to be solved. With the input prompt, ChatGPT can output the desired edited caption with the required format.

\noindent\textbf{In-Context Learning.} Considering the differences between seven multi-grained commands, we manually select seven high-quality input-output demonstrations involving all command types. The command type of given examples will be aligned with the new case to fulfill better results. For example, if we want ChatGPT to edit a video description under the $\langle$\textit{add}, \textit{pos}, \textit{attr}$\rangle$ command, we will give a matched example of $\langle$\textit{add}, \textit{pos}, \textit{attr}$\rangle$ command to help it better solve the task.

\subsection{Instruction-tuning Data of VidLLM}
We convert the original samples from the VATEX-EDIT and EMMAD-EDIT datasets into an instruction-tuning format to facilitate the training of end-to-end video large language models. Figure~\ref{fig:videochat-train-sample} illustrates the converted data sample.

\subsection{Results of GPT4-o }
As Table~\ref{tab:gpt4o} shows, we report the results of GPT-4o\footnote{https://openai.com/index/hello-gpt-4o/} in the challenging EMMAD-EDIT dataset as a reference upper bound of the task. Given that the video interface for GPT-4o is not yet available, we extract 8 frames from each video and assess the model using these multiple images as inputs.

\begin{table}[t]
\centering
\caption{Results of GPT-4o on the EMMAD-EDIT dataset.}
\begin{adjustbox}{max width=\linewidth}

\begin{tabular}{l|ccc}
\toprule
   & Control.       & Fluency       & Text-Vision Align   \\ \midrule
OPA & 66.4  & 76.5  & 44.8  \\
VidLLM (VideoChat) & 62.6 & 81.7 & 45.3 \\
ImgLLM (w/ ChatGPT) & 66.9 & 79.8 & 45.3 \\

GPT-4o & \colorbox{redbackground}{\textcolor{myred}{\textbf{71.1}}} & \colorbox{redbackground}{\textcolor{myred}{\textbf{81.9}}} & \colorbox{redbackground}{\textcolor{myred}{\textbf{45.3}}} \\ \bottomrule
\end{tabular}
\end{adjustbox}

\label{tab:gpt4o}
\end{table}

\begin{figure}[tbp]
    \centering
    \includegraphics[width=\linewidth]{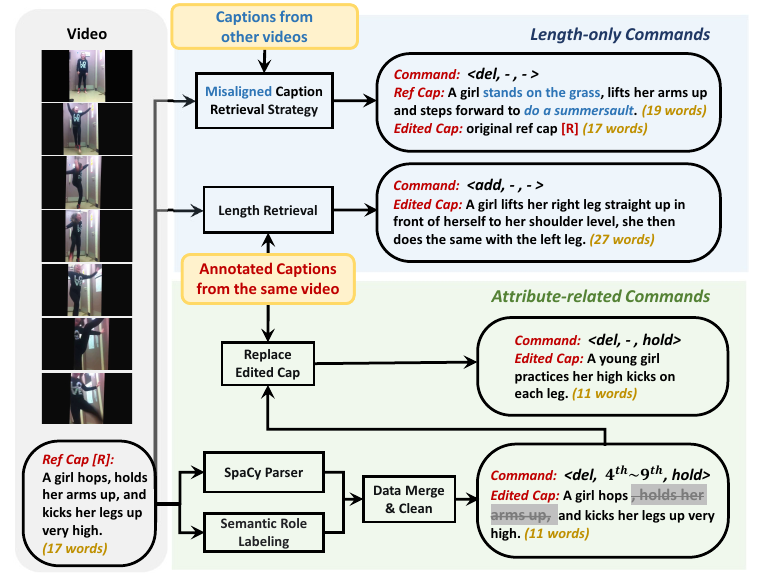}

    \caption{The automatic procedure of VATEX-EDIT dataset construction. The omitted commands, e.g. \textit{$\langle$add, pos, attr$\rangle$ and $\langle$del, pos, -$\rangle$}, can be easily converted from the data samples of shown commands.}
    \label{fig:vatex_data}
\end{figure}

\section{VATEX-EDIT Dataset Details}
\label{sec:vatex-edit}
\subsection{Automatic Dataset Construction}
\label{sec:auto_data}
We construct quadruples \textit{(video, command, reference caption, edited caption)} according to two types of commands including \textbf{coarse-grained length-control commands} and \textbf{fine-grained attribute-related commands}. 
We complement more details about constructing attribute-related commands.

\noindent\textbf{Attribute-related commands.} Our goal is to construct related \textit{(command, edited caption)} samples for each \textit{(video, reference caption)} to support attribute related commands in a ``degradation'' manner. 
The main challenges lie in two aspects: 1) extracting meaningful noun, verb, or modifier attributes in the reference caption $R$ and 2) deleting the attribute-related semantic spans while maintaining the fluency of rest content $R_{\backslash attr}$ to get an attribute-removed caption $Y$. In detail, we adopt four steps as follows:

\begin{itemize}
    \item First, we use the Spacy syntactic dependency parser to build a textual dependency tree that contains the Part-of-Speech information and relationships between tokens. We select reasonable branches in the parsed tree as attributes and further prune the branch to ensure the fluency of the rest caption. 
    \item Second, we use a Semantic Role Labeling model~\cite{shi2019semanticRole} to analyze the semantic roles of each span in a sentence. It can help to better judge whether a parsed attribute span in the first step can be deleted or not, especially noun attributes.
    \item Third, we merge the attributes which only modify one or two tokens to improve the task challenge, that is, the model may be required to edit with multiple attributes in one round.
    \item 
    Finally, considering the intrinsic error of the parsing model, we adopt a post-processing stage to filter low-quality sentences considering sentence fluency, edited token length, and attribute diversity. The ground-truth sentence fluency is further verified in Section~\ref{sec:text_quality}.
\end{itemize}

When the attribute-related spans are removed in a sentence, the edited positions can be naturally recorded.  Through the above steps, we can get high-quality samples for the \textbf{\textit{$\langle$del, pos, attr$\rangle$}}  commands. Meanwhile, reversed samples can be obtained for the \textbf{\textit{$\langle$add, pos, attr$\rangle$}} command by exchanging the reference caption and the edited caption.  For these two fine-grained commands, the sentence length, structure, and semantics are controlled at the same time.

For commands \textbf{\textit{$\langle$add, -, attr$\rangle$}} and \textbf{\textit{$\langle$del, -, attr$\rangle$}} that omit positions, they mainly control the sentence length and semantics, not structure. We get relevant samples by relaxing the structure constraint based on the above ``degradation'' manner. In detail, we replace the edited caption by retrieving desired sentences satisfying both length and semantics (attributes) constraints from original annotated descriptions for the same video.

\subsection{VATEX-EDIT Test set Quality}
\label{sec:text_quality}
\begin{figure}[t]
    \centering
    \includegraphics[width=\linewidth]{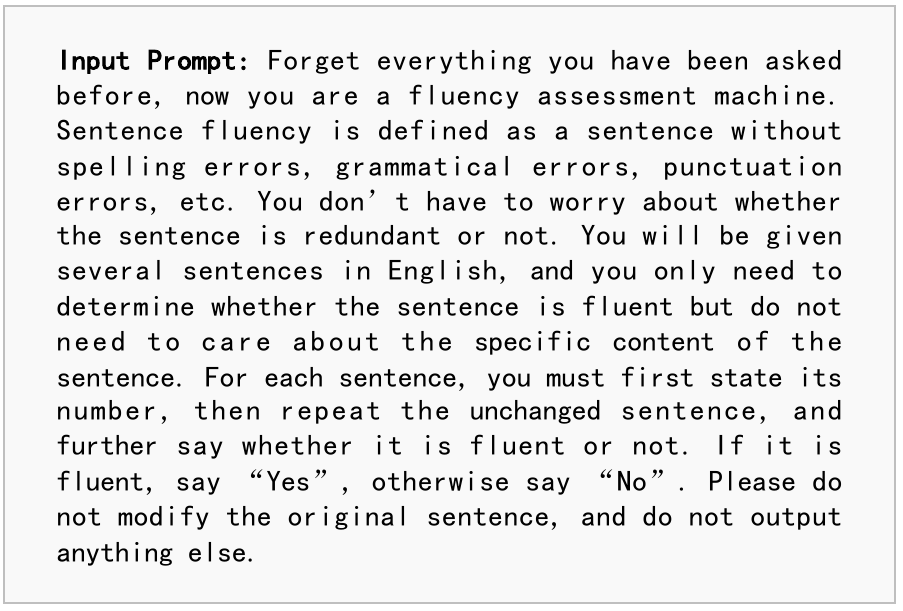}
    \caption{Input prompts of ChatGPT fluency evaluation. }
    \label{fig:prompt}
\end{figure}

In the VATEX-EDIT dataset, it is worth noting that only the ``\textit{$\langle$del, pos, -$\rangle$, shorten description at specified positions}'' command uses the auto-constructed sentences as ground-truth captions in order to control the sentence structure. Meanwhile, the data samples of other commands utilize human-annotated sentences from the original VATEX dataset as ground-truth captions. 
To assess the quality of the auto-constructed sentences, we evaluate the test set of VATEX-EDIT through human and ChatGPT 
evaluations.
For the ChatGPT evaluation, we design suitable prompts (depicted in Figure~\ref{fig:prompt}) to guide ChatGPT to judge whether a sentence is fluent or not. 
We randomly sample 100 ground-truth sentences on the test set and calculate the fluency rate obtained by ChatGPT.
For human evaluation, we recruit 20 crowd workers to rate the fluency score (ranging from 1 to 5) of 200 randomly sampled ground-truth sentences.
As Table~\ref{tab:fluency} shows, the automatically constructed ground-truth sentences of the \textit{``$\langle$del, pos, -$\rangle$''} command are as fluent as the human-annotated ground-truth sentences of other commands (87\% vs 91\% and 4.40 vs 4.37), which indicates the high quality of the VATEX-EDIT test set.

\section{EMMAD-EDIT Dataset Details}
\label{sec:emmad-edit}
We manually collect the high-quality dataset EMMAD-EDIT in the E-commerce domain based on the Chinese E-commerce video captioning dataset E-MMAD. The data sample in E-MMAD dataset consists of a product video, an advertising video description, and additional information including video titles and structure attributes.

It collects 120,984 videos with average duration of 30.4 seconds and the annotated Chinese description length is 67 words on average. The characteristics of \textit{long videos} and \textit{long captions} make it suitable for building a challenging VCE dataset. 
During annotation, we select  data samples from E-MMAD dataset with relatively long descriptions. In addition to videos and  descriptions, we also provide product structure information and video titles for reference.

\begin{table}[t]
\centering
\caption{Fluency evaluation of ground-truth sentences on the VATEX-EDIT test set. ChatGPT measures the fluency (YES/NO) rate of 100 sentences. Human measures the fluency rate (ranging from 1 to 5, 5 is the best) of 200 sentences.}
\begin{tabular}{c|cc} 
\toprule
        & \textbf{Constructed} & \textbf{Annotated} \\ \toprule
\textbf{ChatGPT} & 87\%          & 91\%    \\ \midrule \midrule
\textbf{Human}   & 4.40          & 4.37  \\ \bottomrule
\end{tabular}
\label{tab:fluency}
\end{table}

\subsection{EMMAD-EDIT dataset statistics}

Table~\ref{tab:emmad-edit} shows breakdown data statistics of the EMMAD-EDIT dataset. 
It has overall \textcolor{black}{79,652 editing instances for 16,176 product videos}. Diverse unique attributes and vocabulary also indicate data richness. We separate the \textit{abstract} attribute-related data samples as an extra challenging subset. Considering realistic demands, we utilize all these data samples to construct ``\textit{$\langle$add, -, attr$\rangle$, add attributes in description}'' command cases. Note that in the Experiments section, we present the EMMAD-EDIT results training without the \textit{abstract} subset data by default.


\begin{table*}[t]

\begin{center}
    \caption{Data statistics of EMMAD-EDIT dataset. \textit{VTime} denotes the average time length (seconds) of videos. \textit{$\text{Len}_{Ref}$} denotes the average length of reference captions and \textit{$\text{Len}_{GT}$} is the length of groundtruth captions. \textit{Uni. Attrs} means the vocabulary of annotated attributes. \textit{$\text{Overall}_\textit{Abstract}$} is the challenging subset of abstract attribute-related samples.}
    \vspace{-8pt}
\begin{tabular}{@{}c|rrr|rrr|rrrrrr}
\toprule
  \multirow{2}{*}{\textbf{Command}} &
  \multicolumn{3}{c|}{\textbf{\#Videos}} &
  \multicolumn{3}{c|}{\textbf{\#Editing instances}} &
  \multirow{2}{*}{\textbf{VTime}}&
  \multicolumn{1}{l}{\multirow{2}{*}{\textbf{$\text{Len}_{Ref}$}}} &
  \multicolumn{1}{l}{\multirow{2}{*}{\textbf{$\text{Len}_{GT}$}}} &
  \multicolumn{1}{l}{\multirow{2}{*}{\textbf{Edit Dist}}} &
  \multicolumn{1}{l}{\multirow{2}{*}{\textbf{Uni. Attrs}}} &
  \multicolumn{1}{l}{\multirow{2}{*}{\textbf{Vocab}}} \\ 
 &
  \multicolumn{1}{r}{Train} &
  \multicolumn{1}{r}{Val} &
  \multicolumn{1}{r|}{Test} &
  \multicolumn{1}{r}{Train} &
  \multicolumn{1}{r}{Val} &
  \multicolumn{1}{r|}{Test} &
  \multicolumn{1}{l}{} &
  \multicolumn{1}{l}{} &
  \multicolumn{1}{l}{} &
   \\ \toprule

$\langle$\textit{add},\hfill-\hfill,\hfill-\hfill$\rangle$\hfill        & 7,751 & 2,599 & 2,633 & 7,752  & 2,599 & 2,633 & 26.6 & 72.2  & 100.9 & 29.8 & -     & 29,241 \\
 $\langle$\textit{add}, \textit{pos},\hfill-\hfill$\rangle$\hfill     & 3,221 & 1,077   & 1,094   & 3,221  & 1,077   & 1,094   & 26.3 & 91.6 & 99.4 & 8.7 & -     & 18,640 \\
 $\langle$\textit{add},\hfill-\hfill, \textit{attr}$\rangle$\hfill     & 3,221 & 1,078   & 1,094   & 3,221  &  1,078  & 1,094   & 27.1 & 93.1  & 100.6  & 8.7 & 2,388 & 18,501 \\
 $\langle$\textit{add}, \textit{pos}, \textit{attr}$\rangle$\hfill  & 3,221 & 1,077   & 1,093   & 3,221  & 1,077   & 1,093   & 26.6 & 92.0 & 99.9 & 8.7  & 2,907 & 18,584 \\ 
$\langle$\textit{del},\hfill-\hfill,\hfill-\hfill$\rangle$\hfill       & 7,751 & 2,599 & 2,633 & 7,753  & 2,599 & 2,633 & 26.4 & 100.1  & 71.4  & 29.6 & -     & 29,456 \\
 $\langle$\textit{del}, \textit{pos},\hfill-\hfill$\rangle$\hfill    & 3,221 & 1,078   & 1,095   & 3,221  & 1,078   & 1,095   & 27.1 & 102.1 & 86.1 & 17.2 & -     & 18,734 \\
$\langle$\textit{del}, \hfill-\hfill, \textit{attr}$\rangle$\hfill    & 3,221 & 1,078   & 1,095   & 3,221  & 1,078   & 1,095   & 27.2 & 102.7 & 86.2  & 17.8 & 3,038 & 18,924 \\
$\text{Overall}_\textit{Specific}$                     & 16,176 & 5,418 & 5,502 & 31,610  & 10,586 & 10,737 & 26.9 & 91.3  & 90.4 & 20.8 & 6,003  & 44,725
  \\ \midrule \midrule
$\text{Overall}_\textit{Abstract}$             & 15,955 & 5,328 & 5,432 & 15,959  & 5,328 & 5,432 & 26.8 & 90.9  & 100.9 & 11.4 & 648  & 44,347
  \\  
  \bottomrule
\end{tabular}
\label{tab:emmad-edit}
\end{center}
\end{table*}

\subsection{The Impact of Data Volume.}
\begin{figure}[H]
    \centering
    \includegraphics[width=\linewidth]{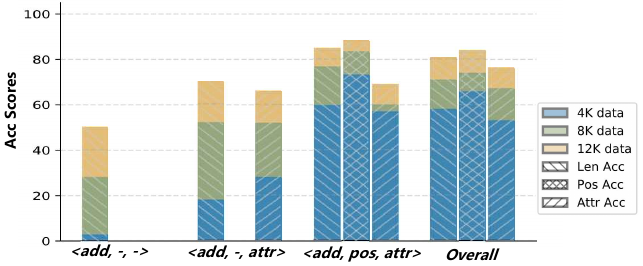}
    \caption{Overall and breakdown performance under different data volumes (4K/8K/12K data samples).}
    \label{fig:data_ratio_overall}
\end{figure}

Considering the limited scale of manually collected data in EMMAD-EDIT, we analyze the results under different volume data with 4K, 8K, 12K samples. Figure~\ref{fig:data_ratio_overall} shows that a growing volume of data consistently increases the controllable scores. Breakdown analysis of multi-grained commands reveals that more challenging commands, e.g. \textit{$\langle$add,-,-$\rangle$}, require higher volume of training data samples to get desired performance.

\begin{figure}[H]
    \centering
    \includegraphics[width=\linewidth]{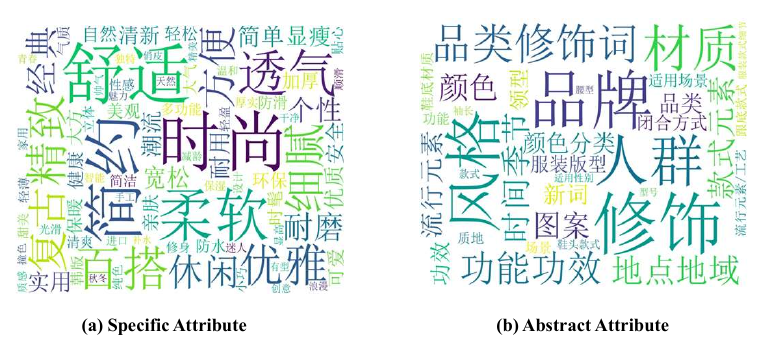}
    \caption{Word clouds of attributes on the EMMAD-EDIT dataset. The left shows the diversity of annotated specific attributes and the right shows the abstract attributes.}
    \label{fig:emmad_attr}
\end{figure}

\subsection{Data Visualization}

We manually collect data samples that support the editing of two types of attributes: \textit{specific} and \textit{abstract}.  \textit{Specific} attributes directly appear in the reference caption to support straightforward content editing such as ``comfortable'' and ``fashion''. Meantime, \textit{abstract} attributes are more high-level concepts that may consist of multiple specific attributes. For example, ``Style'', ``Target People'', and ``Time and Seasons'' are annotated \textit{abstract} attributes in the dataset.
The word clouds (shown in Figure~\ref{fig:emmad_attr}) of \textit{specific} and \textit{abstract} attributes show the diversity of the EMMAD-EDIT dataset.

\begin{figure*}[t]
    \centering
    \includegraphics[width=0.9\linewidth]{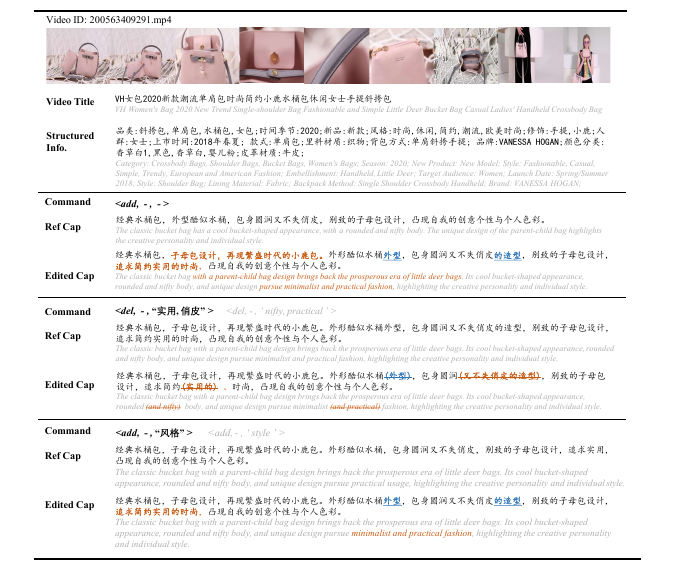}
    \vspace{-8pt}
    \caption{Data samples of annotated EMMAD-EDIT dataset.}
    \label{fig:emmad_samples}
\end{figure*}

\begin{figure*}[t]
    \centering
    \includegraphics[width=0.95\linewidth]{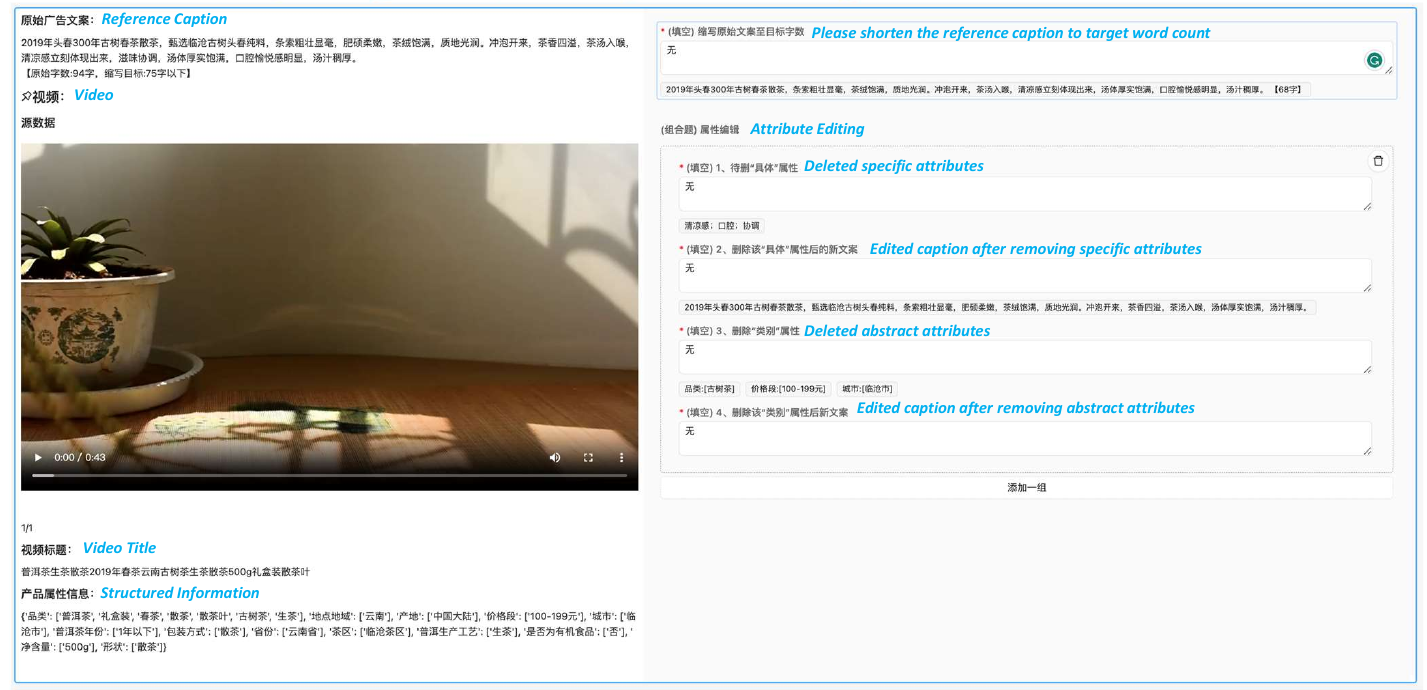}
    \vspace{-8pt}
    \caption{Annotation interface for constructing the EMMAD-EDIT dataset.}
    \label{fig:emmad_web}
\end{figure*}

We visualize annotated samples of the EMMAD-EDIT dataset in Figure~\ref{fig:emmad_samples}. 
Besides quadruples \textit{(video, command, reference caption, edited caption)} that support video caption editing, we also provide additional product information such as structured information and the video title.
The annotation interface during data construction is shown in Figure~\ref{fig:emmad_web}.

\section{Metric Details}
\label{sec:metric}

To support evaluations in Chinese, we utilize a Chinese GPT-2~\cite{GPT2-Chinese} to calculate Chinese sentence likelihood and then obtain PPL scores. For EMScore, we replace the core vision-language alignment model EN-CLIP~\cite{clip} with CN-CLIP~\cite{chinese-clip}.

\subsection{Position Accuracy Design}
\label{sec:pos_acc}

We propose a novel \textit{Position Accuracy (Pos-Acc)} metric to measure whether models insert/remove the content in the specified positions under fine-grained editing control. 
The challenge of calculating Pos-Acc is the misalignment between the  reference caption $R$ and the edited caption $Y$. The two textual sentences have variable lengths and the edited operation may appear in the other positions to maintain the overall fluency.
To tackle the above challenge, we propose a novel Dynamic Sequence Aligning (DSA) algorithm to align two variable-length textual sequences based on the absolute positions, inspired by classical Dynamic Time Warping (DTW)~\cite{muller2007dtw}. 

The pseudo-code is presented in Figure~\ref{fig:torch_implementation}, which can align two variable-length text sequences in positional indexes, resulting in related spans \{$S_{m_1}, S_{m_2},\dots, S_{m_K}$\} in $Y$ aligned to [MASK] tokens $\{m_1, m_2,\dots, m_K\}$ in $R$. We count the percentage of correct samples that insert/remove new content $S_{m_K}$ in given position $m_K$.

\begin{figure*}[t]
    \centering
    \includegraphics[width=0.95\linewidth]{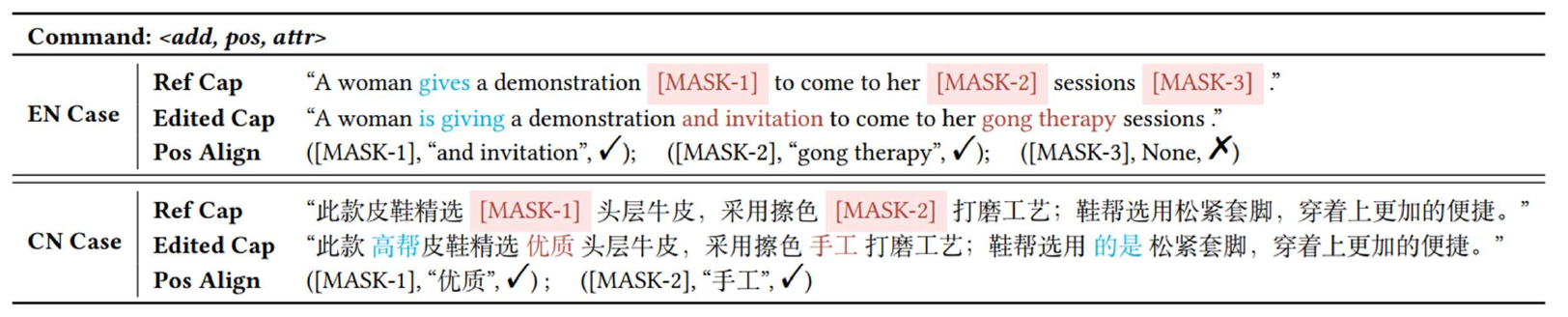}
    \caption{Cases of aligned reference captions and edited captions by the proposed Dynamic Sequence Aligning Algorithm.}
    \label{fig:pos_case}
\end{figure*}

As Figure~\ref{fig:pos_case} illustrates, we visualize two aligned cases in English and Chinese respectively obtained by the DSA algorithm.

\begin{figure}[hbtp]
\begin{lstlisting}[language=Python]
# R is the word sequence of Reference Caption
# Y is the word sequence of Edited Caption
def DynamicSeqAlign(R, Y):
    # Initialize DTW distance matrix
    DTW = {}
    for i in range(len(R)):
        DTW[(i, -1)] = float('inf')
    for j in range(len(Y)):
        DTW[(-1, j)] = float('inf')
    DTW[(-1, -1)] = 0
    # Initialize the aligned path matrix
    PATH = {}
    # Dynamic programming to align R and Y
    for i in range(len(R)):
        for j in range(len(Y)):
            # get distance between word i and word j
            dist = get_dist(R[i], Y[j])
            # get the min distance from last time step
            min_dist = min(DTW[(i-1, j)], DTW[(i, j-1)], DTW[(i-1, j-1)])
            # update DTW matrix for current time step 
            DTW[(i, j)] = dist + min_dist
            # record the related distance path
            PATH = update(PATH)
    # get the minimum distance path aligning R and Y
    best_path = trace(PATH)
    # filter repetitive words in Y
    align_path = filter(best_path)
    return align_path

# get distance between two words from R and Y
def get_dist(R_word, Y_word):
    if R_word == Y_word:        # same words
        return 0
    elif R_word == '[MASK]':    # ([MASK], Y_word)
        return 100
    else:                       # unmatched words
        return float('inf')
\end{lstlisting}
\caption{Pseudo-code for Dynamic Sequence Aligning in a Python-like style.} 
\label{fig:torch_implementation}
\end{figure}

\section{Conversion from Interface Signals to Triplet Format}
\label{sec:conversion}
Though the main focus of this paper lies in exploring the novel VCE task utilizing triplet commands, we also shed light on how to convert the prevalent interface signals, i.e. natural language and handwriting editing trajectories, to the triplet formatted controls.

\subsection{From Natural Language to Triplets}

To convert natural language signals, we can adopt fuzzy matching to recognize add or delete operations and use text parser tools to get specific semantic roles of a sentence. We can also leverage the ability of LLMs such as ChatGPT and LLaMA-2, which already have outstanding language understanding and summarization capabilities. Specifically, we can design suitable prompts to convert text sentences into triplets to meet the pre-defined requirements. 

\subsection{From Triplets to Natural Language}
The triplet command in our annotated datasets can be easily converted to natural languages to support more diverse scenarios and applications. For example, using natural language to explore the capability of  LLMs in the VCE task (\ref{sec:chatgpt}). Specifically, we design a series of templates shown in Table~\ref{tab:triplet2sent} to convert the triplet command to natural language. We will also release the two benchmark datasets with user commands both in the triplet and natural language format to benefit the community.

\subsection{From Handwriting-revision Traces to Triplets}
\begin{figure*}[t]
    \centering
    \includegraphics[width=0.9\linewidth]{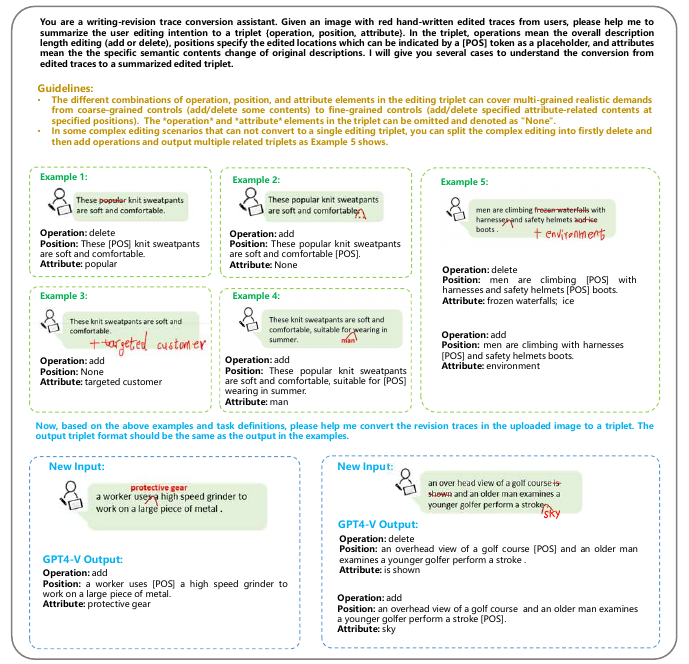}
    \caption{Designed prompts and output instances of GPT-4V to convert the human-writing editing trace as an image to the triplet format control.}
    \label{fig:gpt-4v}
\end{figure*}

The recent advancement of GPT-4Vision (GPT-4V) has shown its powerful capability of Visual Referring Prompting~\cite{yang2023dawn}. GPT-4V can well understand visual pointers (such as circles, arrows or traces) directly drawn on images, therefore, revealing a novel human-model interaction method called ``visual referring prompting''. Combined with its accurate OCR capability, GPT-4V can serve as an ideal tool to input the handwriting editing traces as an image and convert it to the triplet control output, as illustrated in Figure~\ref{fig:gpt-4v}.

\section{Future Directions}
\label{sec:future}
In this paper, we make the first attempt to propose the novel Video Caption Editing (VCE) task and collect two benchmark datasets to support it. We further propose a unified framework OPA for VCE and compare it with a ChatGPT pipeline. Based on the task, dataset and method foundation, we aim to make a good start for the community. There are various interesting aspects worth exploring in the future:
\begin{itemize}
    \item \textbf{A versatile system for video captioning and editing. } The dense annotation of the quadruple \textit{(video, command, reference caption, edited caption)} in our dataset has the potential to support building a versatile system encompassing conventional video captioning, controllable video caption and video caption editing. For initialization, conventional video captioning can produce a generated description for a given video. subsequently, video caption editing can be utilized to update and revise the original description. When omitting the reference caption, the rest annotation \textit{(video, command, edited caption)} can be adjusted to achieve controllable video captioning.
    \item \textbf{A more robust system for poor reference caption.} In the VCE task, the reference caption can be the edited caption from the last round to fulfill multi-round editing. Furthermore, it can also be extended with human-written drafts or machine-generated captions. Under these circumstances, the robustness of the VCE system to low-quality reference captions should be taken into consideration. 
    \item \textbf{The abstract-attribute subset of EMMAD-EDIT is under-explored.} The abstract-attribute subset (details in Sec 4.2.) involves abstract attributes that do not directly appear in the reference caption. It requires models to understand and reason the video content at a higher semantic level. In the experiments, we only assess the performances of OPA and ChatGPT pipeline on the easier \textit{specific} subset, whose performances are not yet so desirable, the \textit{abstract} subset therefore remains a challenge for future exploration.
    \item \textbf{Serve as a touchstone for video large language models (VidLLMs).} The recent emergence of VidLLMs such as GPT4-Vision~\cite{gpt4v}, VideoChat~\cite{videochat}, and VideoChatGPT~\cite{Maaz2023VideoChatGPTTD} has ignited sparks for a generalist video assistant. However, existing research also points out their limitations for long video understanding~\cite{jin2023chatunivi}, visual hallucination~\cite{yin2023woodpecker} and multi-modal instruction following capability~\cite{ye2023mplugowl2}. The VCE benchmark can serve as a new multi-modal evaluation for assessing both long video understanding and multi-grained text editing abilities.
    
\end{itemize}

\begin{table*}[t]
\centering
\caption{Defined templates that conveniently convert triplet commands to natural language format. The '\{\}' represents the placeholder of specific attributes.}
\begin{tabular}{c|l}

\toprule
\textbf{Command}   & \qquad \qquad \qquad \qquad \qquad \textbf{Conversion Template}                                   \\ \toprule
$\langle$\textit{del},\hfill-\hfill,\hfill \textit{attr} \hfill$\rangle$\hfill & Delete contents about '\{\}' from the reference description.  \\
 $\langle$\textit{add}, \textit{pos}, \textit{attr}$\rangle$\hfill  & Add contents about '\{\}' at {[}POS{]}.                        \\
$\langle$\textit{add},\hfill-\hfill, \textit{attr}$\rangle$\hfill   & Add contents about '\{\}' to expand the reference description. \\
 $\langle$\textit{add}, \textit{pos},\hfill-\hfill$\rangle$\hfill   & Add video-related contents at {[}POS{]}.                       \\
$\langle$\textit{del}, \textit{pos},\hfill-\hfill$\rangle$\hfill  & Contents have been deleted at {[}POS{]}, please make the rest sentence fluent and coherent. \\
$\langle$\textit{del},\hfill-\hfill,\hfill-\hfill$\rangle$\hfill    & Shorten the reference description.                             \\
$\langle$\textit{del}, \hfill-\hfill, \textit{attr}$\rangle$\hfill     & Expand the reference description.    \\ \bottomrule                
\end{tabular}

\label{tab:triplet2sent}
\end{table*}

\section{Social Impacts}
\label{sec:social}
The proposed Video Caption Editing (VCE) task addresses significant challenges in the realm of video content management, offering far-reaching social implications.

\noindent \textbf{Enhanced Accessibility.} By allowing for multi-grained user control over video captions, our system can significantly improve accessibility for individuals with disabilities. Users who rely on captions for understanding video content, such as the hearing-impaired, can benefit from more precise and customizable descriptions tailored to their specific needs.

\noindent \textbf{E-commerce and Marketing.} The integration of our VCE system into e-commerce platforms can revolutionize how product videos are presented. By enabling dynamic and detailed video descriptions that cater to individual consumer preferences, businesses can enhance user engagement and improve product understanding.

\noindent \textbf{User Empowerment and Creativity.} The ability to edit video captions interactively empowers users to tailor content to their specific and creative desires. This democratizes content creation, allowing individuals and small creators to produce high-quality, customized video content without requiring extensive technical expertise.










\end{document}